\documentclass{article}

 \usepackage[preprint]{neurips_2026}

\usepackage[utf8]{inputenc} 
\usepackage[T1]{fontenc}    
\usepackage{hyperref}       
\usepackage{url}            
\usepackage{booktabs}       
\usepackage{amsfonts}       
\usepackage{nicefrac}       
\usepackage{microtype}      
\usepackage{xcolor}         

\usepackage{amsmath}
\usepackage{amssymb}
\usepackage{mathtools}
\usepackage{amsthm}
\usepackage{booktabs}
\usepackage{multirow}
\usepackage{tikz} 
\usepackage{microtype}
\usepackage{graphicx}
\usepackage{subcaption}
\usepackage{bbm}
\usepackage{algorithm}
\usepackage{algpseudocode}

\newcommand{\model}{MO-ARM}

\def\gL{{\mathcal{L}}}

\def\E{{\mathbb{E}}}

\def\x{{\bm{x}}}
\def\xo{{\bm{x}_{\bm{z}_\o}}}
\def\xm{{\bm{x}_{\bm{z}_\m}}}

\def\z{{\bm{z}}}
\def\zo{{\bm{z}_\o}}
\def\zm{{\bm{z}_\m}}

\def\Lo{{L_\o}}
\def\Lm{{L_m}}

\def\o{{\mathrm{o}}}
\def\m{{\mathrm{m}}}

\definecolor{darkgrey}{rgb}{0, 0.306, 0.678}

\usepackage{amsmath,amsfonts,bm}

\def\eqref#1{equation~\ref{#1}}

\def\1{\bm{1}}

\DeclareMathAlphabet{\mathsfit}{\encodingdefault}{\sfdefault}{m}{sl}
\SetMathAlphabet{\mathsfit}{bold}{\encodingdefault}{\sfdefault}{bx}{n}

\def\gL{{\mathcal{L}}}

\title{Order-Agnostic Autoregressive Modelling \\
	with Missing Data}

\author{%
	Ignacio Peis \\
	Technical University of Denmark \& \\
	Pioneer Centre for AI \\
	\texttt{ipeaz@dtu.dk} \\
	\AND
	Pablo M.~Olmos  \\
	Universidad Carlos III de Madrid \\
	\texttt{olmos@tsc.uc3m.es} \\
	\And
	Jes Frellsen \\
	Technical University of Denmark \& \\
	Pioneer Centre for AI \\
	\texttt{jefr@dtu.dk} \\
}

\begin{document}

\maketitle

\begin{abstract}
Order-Agnostic autoregressive models have demonstrated strong performance in deep generative modeling, yet their use in settings with incomplete data remains largely unexplored. In this work, we reinterpret them through the lens of missing data. First, we show that their standard training procedure on fully observed data implicitly performs imputation under a missing completely at random mechanism, resulting in robust out-of-sample imputation performance in settings with high missingness. Second, we introduce the first principled framework for training them directly on incomplete datasets under general missingness mechanisms. Third, we leverage their amortized conditional density estimation to perform active information acquisition, i.e., sequentially selecting the most informative missing variables for downstream prediction or inference. Across a suite of real-world benchmarks, our Missingness-Aware Order-Agnostic Autoregressive Model (\model) consistently outperforms established imputation baselines.\end{abstract}

\section{Introduction}

Real-world learning systems frequently operate under partial observability: features may be missing due to sensor faults, acquisition costs, privacy constraints, or design choices in data collection. Models deployed in such environments usually need to (i) learn from incomplete data, (ii) impute missing values at inference time, and (iii) decide which unobserved variables to acquire when observations are costly \citep{little2019statistical, ma2019eddi}. These challenges arise in domains such as medical diagnostics \citep{lin2008exploiting, sterne2009multiple}, scientific experimentation \citep{imhof2004optimal, jarrett1978analysis}, or online decision making \cite{melville2004active, saar2009active}, where accurate conditional reasoning under arbitrary missingness patterns is essential.

Deep generative models offer, in principle, a unified way to reason about incomplete data: by modeling the full joint distribution, one could derive all conditional distributions needed for learning from partial observations, imputing missing values, and selecting which variables to acquire. However, most deep generative models do not make these conditionals tractable or reliable in practice. Variational Autoencoders \citep{kingma2013auto} and their extensions to handle missing data \citep{ma2020vaem, mattei2019miwae, nazabal2020handling, peis2022missing} rely on approximate inference, which limits conditional accuracy and complicates their use for sequential acquisition. Other approaches based on Generative Adversaral Networks \cite{NIPS2014_f033ed80}, such as GAIN \citep{yoon2018gain}, produce plausible imputations but lack likelihoods and principled conditional densities, preventing uncertainty-based decision making. Denoising diffusion models \citep{ho2020denoising} generate high-quality samples but require long iterative sampling chains and struggle to condition precisely on arbitrary subsets of observed variables \citep{lugmayr2022repaint}.\looseness=-1

Autoregressive models, by contrast, provide tractable likelihoods and conditional sampling through their factorized structure. However, classical autoregressive models rely on a fixed generation order, which prevents conditioning on arbitrary subsets of variables. Order-Agnostic Autoregressive Models (OA-ARMs) address this limitation by training on random permutations of the variables \citep{hoogeboom2022autoregressive, uria2016neural, uria2014deep}, effectively treating the generation order as an implicit latent variable. These models amortize over all possible orders through a single neural network, with the goal of learning representations that remain consistent across many sampled permutations. Variants that explicitly learn an order distribution via variational inference \citep{wanglearning} further reinforce this latent-variable interpretation. Computing exact conditionals would require marginalizing over all orders consistent with the observed subset, which is not practical given the $L!$ possible permutations for $L$ variables. In practice, OA-ARMs yield highly accurate approximations to these conditionals, which enables flexible inference while retaining the tractable likelihoods of autoregressive models. Despite this flexibility, their potential for missing-data applications has remained largely unexplored: OA-ARMs have been primarily developed for density estimation rather than for learning, imputing, or reasoning under incomplete data.

In this work, we revisit OA-ARMs through the lens of missingness and uncover an alignment between their training objective and the requirements of incomplete data inference. We show that standard OA-ARM training on fully observed data implicitly simulates the imputation of Missing Completely At Random (MCAR) patterns, yielding strong out-of-sample imputation performance without any modification. Building on this insight, we introduce the first principled framework for training OA-ARMs directly from incomplete datasets under general missingness mechanisms, spanning MCAR (the simplest setting) to MNAR (the most challenging), while preserving exact likelihoods and supporting efficient conditional sampling. Beyond imputation, many practical decision-making scenarios require models not only to fill in missing values but also to determine \emph{which} unobserved variables would be most valuable to acquire. OA-ARMs are uniquely well-suited to this setting because their amortized conditional distributions estimate, for any observed subset, how uncertainty would contract or predictive utility would increase after revealing additional variables. We leverage this property to perform \emph{active information acquisition}, sequentially selecting unobserved variables that maximize downstream predictive performance.

We refer to this extended framework as \emph{Missingness-Aware Order-Agnostic Autoregressive Models} (\model). Across a suite of real-world benchmarks encompassing varied data types such as images and tabular data, \model~ consistently outperforms established generative methods for imputation and active acquisition, demonstrating that OA-ARMs constitute a powerful and previously underappreciated tool for reasoning under incomplete data.

\section{Background} \label{sec:background}

\subsection{Notation}

We denote a data point by $\x \in \mathcal{X}$, where $\mathcal{X} \subseteq \mathbb{R}^D$ for tabular data and $\mathcal{X} = \mathbb{R}^{C \times H \times W}$ for images. We index each atomic element of $\x$ by $j \in [L]$, where an atomic element is a feature in tabular data or a pixel in an image. We set $L := D$ for tabular inputs and $L := H \times W$ for images. 

We use $z_i$ to denote indices in $[L]$, i.e., categorical random variables defined on a context-dependent candidate set of currently available elements. We denote by $\z_{<i}$ the cumulative set of indices selected before step $i$, i.e., $\z_{<i} = \{z_1, \dots, z_{i-1}\}$. For any subset $A \subseteq [L]$, we write $\x_A$ for the subvector (or subimage) formed by the elements of $\x$ indexed by $A$. When convenient, especially in Section~\ref{sec:moarm_prob_model}, we represent such cumulative index sets through their binary-mask encoding. To describe missing data, we write $\x = (\xo, \xm)$, where $\xo$ are the observed entries and $\xm$ are the missing ones, corresponding to the complementary index sets $\zo$ and $\zm = [L] \setminus \zo$, respectively. 



\subsection{Order-Agnostic Autoregressive Models}

Classical ARMs impose a single fixed ordering of the $L$ variables. In contrast, OA-ARMs \citep{uria2016neural, uria2014deep} remove this restriction by defining an autoregressive factorization for any random permutation of the order. More recently, Autoregressive Diffusion Models (ARDMs \cite{hoogeboom2022autoregressive}) proposed an instantiation of OA-ARMs where a neural backbone amortizes the conditional distributions across all permutations by applying appropriate masking, enabling efficient computation. 

Instead of modeling permutations directly, OA-ARMs can be reformulated in terms of a sequence of $L$ \emph{unmasking} latent variables, $\{z_1,\dots,z_L\}$ \citep{wanglearning}, defining an order through a Markov chain:
\begin{equation} \label{eq:oarm_masks}
    p(\z) = \prod_{i=1}^L p(z_i \mid \z_{<i}),
\end{equation}

where $\z_{<1} = \emptyset$, $p(z_1)$ is uniform over $\{1,\dots,L\}$, and each $p(z_i \mid \z_{<i})$ is uniform over the remaining unselected indices. 
During generation, these indices are accumulated to form $\z_{< i}$. This construction  yields a distribution over orders equivalent to uniform permutations.

Figure~\ref{fig:graph_oarm} illustrates this construction. Under this formulation, the conditional likelihood becomes
\begin{equation} \label{eq:oarm_factorization}
    p_\theta(\x|\z) = \prod_{i=1}^L p_\theta(\x_{z_i} \mid \x_{\z_{<i}}).
\end{equation}
OA-ARMs can then be viewed as discrete latent-variable models in which unmasking trajectories act as latent variables generating samples, with joint $p_\theta(\x, \z) = p(\z)\, p_\theta(\x \mid \z)$ and marginal $p_\theta(\x) = \E_{p(\z)}[p_\theta(\x \mid \z)]$. Applying Jensen's inequality yields the Evidence Lower Bound:
\begin{equation} \label{eq:elbo_oarm_act}
    \begin{aligned}
        \log p_\theta(\x) \geq 
        \E_{p(\z)}\left[ \log p_\theta(\x \mid \z) \right] = 
        \sum_\z \sum_{i=1}^L p_\theta(z_i \mid \z_{<i}) \,  \log p(\x_{z_i} \mid \x_{\z_{<i}}) = \gL(\x, \theta).
    \end{aligned}
\end{equation}
Recent work \cite{wanglearning} proposes learning a variational posterior $q_\theta(\z | \x)$ over orders, enabling data-dependent orderings (LO-ARMs). In this work, however, we adopt the simpler and computationally efficient choice $q(\z | \x) = p(\z)$ and assume uniform categorical distributions, which suffices for our setting of learning under missing data.

\subsection{Deep generative models for missing data} \label{sec:dgms_missing}
Deep generative models can naturally handle missing data by treating the unobserved components as latent variables and integrating them out of the joint distribution:
\begin{equation}\label{eq:ll_obs}
    p_\theta(\xo) = \int p_\theta(\xo, \xm)\, d\xm,
\end{equation}
where learning proceeds by maximizing the observed-data log-likelihood. Variational Autoencoder (VAE) \citep{kingma2013auto} extensions such as MIWAE~\citep{mattei2019miwae}, HI-VAE~\citep{nazabal2020handling}, VAEM~\citep{ma2020vaem}, and HH-VAEM~\citep{peis2022missing} approximate this integral via variational inference, assuming conditional independence between $\xo$ and $\xm$ given latent variables. These models have been shown to produce robust imputations under MCAR and MAR missingness mechanisms.

However, as noted in~\cite{ipsennot}, maximizing~\eqref{eq:ll_obs} leads to consistent estimation only when the missingness mechanism is independent of the unobserved values. To formalize this, one can include the observation mask $\zo$ as part of the inference process:
\begin{equation}\label{eq:ll_xo_s}
    p_{\theta,\phi}(\xo, \zo )
    = \int p_\theta(\xo, \xm)\, p_\phi(\zo \mid \xo, \xm)\, d\xm.
\end{equation}
Under MCAR or MAR~\citep{little2019statistical}, \(p(\zo \mid  \xo, \xm)\) simplifies to being independent of $\xm$, and \eqref{eq:ll_obs} remains valid. In contrast, under MNAR (Missing Not At Random), missingness depends on \(\xm\), making the marginal likelihood \(p_\theta(\xo)\) insufficient for unbiased learning. Importantly, many real-world missingness processes are of such type: sensor failures, equipment malfunctions, thresholding effects, or content-dependent corruption all depend on the unobserved signal itself. 

To address MNAR, models like not-MIWAE~\citep{ipsennot} explicitly learn the missingness process using importance-weighted bounds to estimate the joint likelihood in~\eqref{eq:ll_xo_s}. Similarly, adversarial models such as MisGAN~\citep{li2018learning} jointly learn data and mask distributions, enabling imputation under MNAR assumptions.

Beyond VAEs, other deep generative approaches have also been adapted. GAN-based \citep{goodfellow2014generative} methods include GAIN~\citep{yoon2018gain}, which uses a hint mechanism and adversarial training for MAR scenarios, and PBiGAN~\citep{li2020learning}, which extends bidirectional GANs to incomplete inputs. Normalizing Flows \citep{papamakarios2021normalizing} have been applied via MCFlow~\citep{richardson2020mcflow}, which uses Monte Carlo integration to marginalize missing values. Diffusion-based models such as CSDI~\citep{tashiro2021csdi}, NewImp~\citep{chen2024rethinking} or DiffPuter \citep{zhang2025diffputer} treat imputation as conditional generation, achieving state-of-the-art results by iteratively refining samples conditioned on $\xo$.\looseness=-1
\section{Missingness-Aware Order-agnostic Autoregressive Models} \label{sec:method}

This section presents our methodological contributions. We begin in Section~\ref{sec:preliminaries} by introducing theoretical results that refine the understanding of OA-ARMs through the lens of missing data. Sections~\ref{sec:moarm_prob_model}--\ref{sec:imputation} then describe the components and capabilities of our proposed model, \model{}, and Section~\ref{sec:saia} details how it enables effective active information acquisition.

\subsection{Preliminaries} \label{sec:preliminaries}

We now show that the OA-ARM training objective derived from the Markovian formulation induces a training-time corruption process that is equivalent to simulating MCAR masks. This connection provides intuition for why OA-ARMs naturally excel at imputation even when trained exclusively on fully observed data.

Starting from the Markovian ELBO in \eqref{eq:elbo_oarm_act}, we decompose it into $L$ contributions, each corresponding to one unmasking step:
\begin{equation}
    \gL (\theta, \x) = \sum_{i=1}^L \sum_{\z_{\leq i}} p(z_i \mid \z_{<i}) \log p_\theta(\x_{z_i} \mid \x_{\z_{<i}}) = 
    \sum_{i=1}^L \gL_i(\x, \theta),
\end{equation}
where each $\gL_i$ describes the likelihood of a new variable $\x_{z_i}$ from a partially revealed subset $\x_{\z_{<i}}$.

Computing all $\gL_i$ terms is computationally expensive for high-dimensional data. Following the discrete-time diffusion literature \citep{ho2020denoising, songdenoising}, we apply the Law of the Unconscious Statistician (LOTUS):
\begin{equation} \label{eq:oaarm_objective_lotus}
    \gL (\theta, \x) = L \cdot \E_{i\sim U(1,L)} \left[ \gL_i(\x, \theta) \right],
\end{equation}
which yields an unbiased estimator by sampling a single step index $i$.

Crucially, under the uniform prior, every element is equally likely
to occupy any of the $L$ autoregressive positions. Thus, the probability that element $j$
has already appeared by step $i-1$ is
$p(j \in \z_{<i}) = (i-1)/L$, which allows us to easily sample from the marginals. Since
$p(\z_{\leq i}) = p(\z_{<i})\,p(z_i \mid \z_{<i})$, we can estimate each
$\gL_i$ using one sample $\z_{<i} \sim p(\z_{<i})$:
\begin{equation}
    \hat{\gL}_i (\theta, \x)
    = \sum_{\z_{i}} p(z_i \mid \z_{<i}) \, \log p_\theta(\x_{z_i} \mid \x_{\z_{<i}}) 
    = \frac{1}{L-i+1} \sum_{z_i \in \z{\geq i}} \log p_\theta(\x_{z_i} \mid \x_{\z_{<i}}),
\end{equation}
under the uniform Categorical assumption for $p(z_i | \z_{<i})$.
Substituting into \eqref{eq:oaarm_objective_lotus} yields the final OA-ARM objective:
\begin{equation} \label{eq:oarm_obj}
    \hat{\gL} (\theta, \x) =  \frac{L}{L-i+1} \sum_{z_i \in \z_{\geq i}} \log p_\theta(\x_{z_i} \mid \x_{\z_{<i}}),
\end{equation}
with $i \sim U(1,L)$ and $\z_{<i} \sim p(\z_{<i})$.

\paragraph{Connection to MCAR missingness.}
Note that $\z_{<i}$ is distributed exactly as a uniformly selected subset of size $i-1$ from $\{1,\dots,L\}$, so every variable has the same probability of being masked or unmasked at each step, independently of its value. Consequently, each training iteration asks the model to predict a randomly masked subset of the data from its complementary observed part, which precisely mirrors MCAR imputation. This explains why OA-ARMs, although trained only on fully observed data, naturally generalize to missing-data settings and achieve strong out-of-sample performance under MCAR missingness.

\subsection{MO-ARM Probabilistic Model} \label{sec:moarm_prob_model}

\begin{figure}[t]
    \centering
    \begin{subfigure}[t]{0.48\linewidth}
        \centering
        \includegraphics[height=0.72\linewidth]{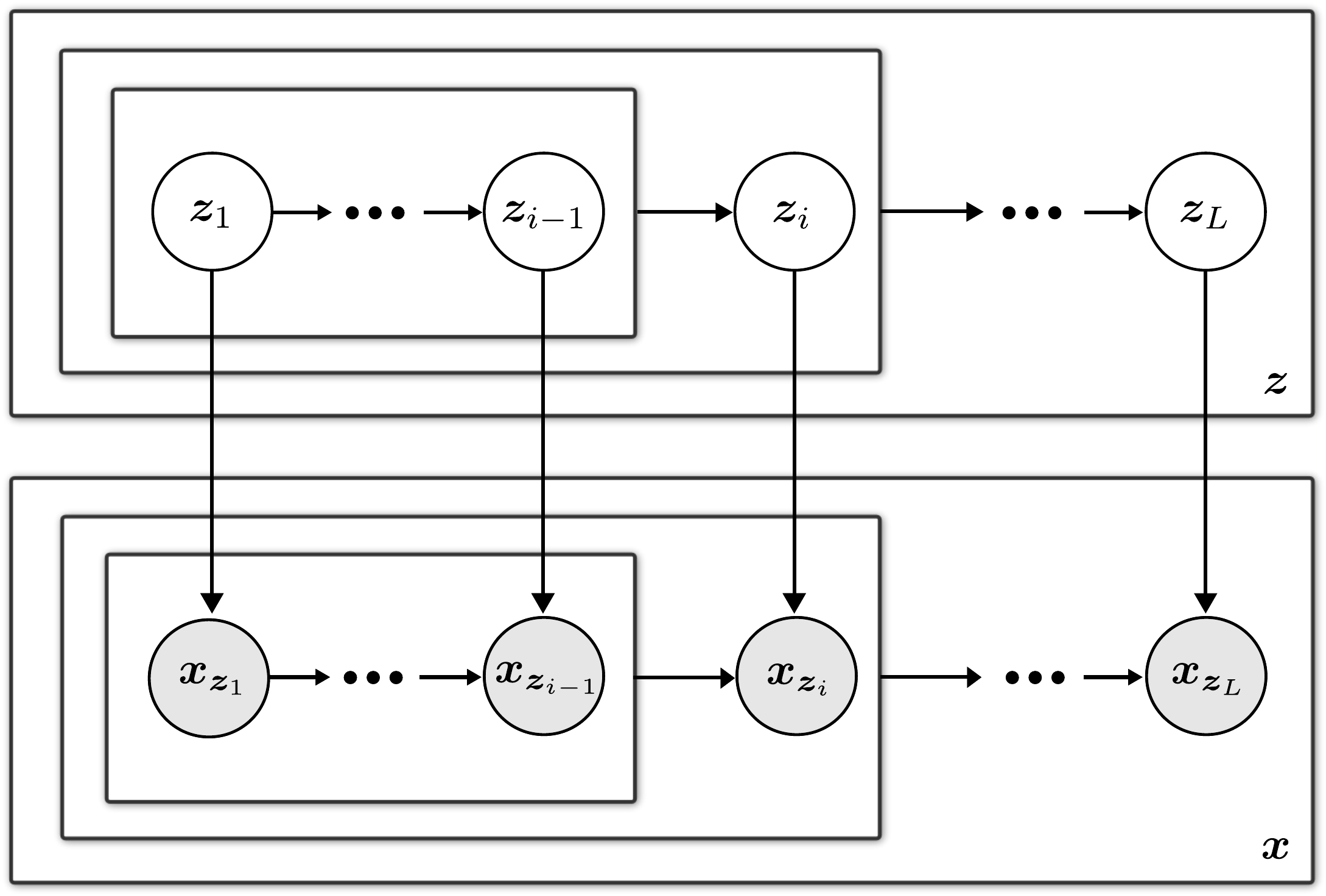}
        \caption{OA-ARM graphical model.}
        \label{fig:graph_oarm}
    \end{subfigure}
    \hfill
    \begin{subfigure}[t]{0.48\linewidth}
        \centering
        \includegraphics[height=0.72\linewidth]{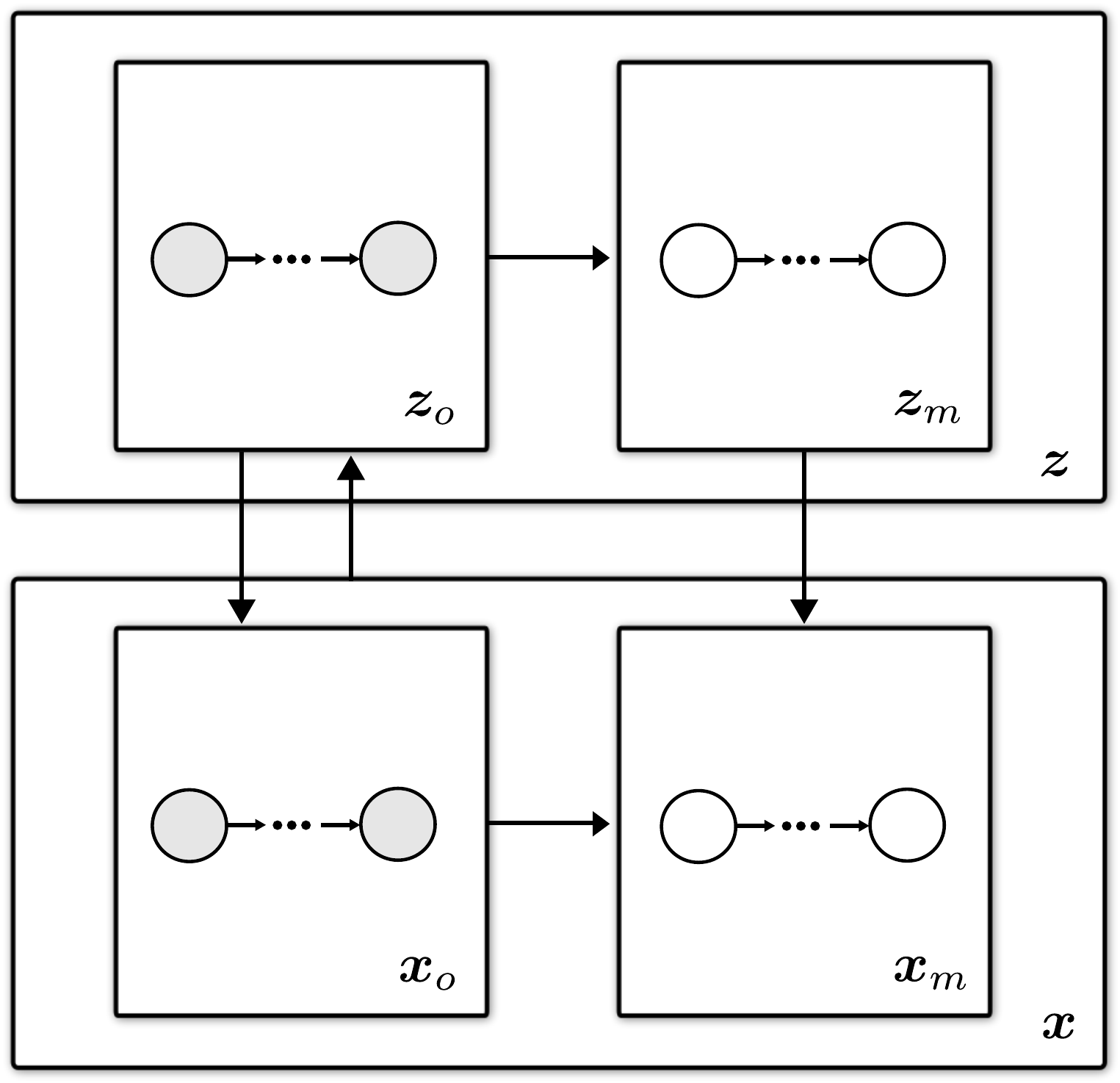}
        \caption{MO-ARM inference model.}
        \label{fig:graph_moarm_inference}
    \end{subfigure}
    \caption{Graphical models used throughout the paper. Boxes denote collections of variables. The left panel illustrates the OA-ARM formulation, where latent variables $\z = (z_1,\dots,z_L)$ define a permutation over the $L$ elements and observed variables $\x = (\x_{z_1},\dots,\x_{z_L})$ are generated autoregressively. The right panel shows the missingness-aware inference model used for imputation.}
    \label{fig:graphical_model}
\end{figure}

We now describe the probabilistic model underlying \model{}, illustrated in
Figure~\ref{fig:graph_moarm_inference}. Our goal is to model incomplete data
$(\xo, \xm)$, where $\xm$ is a latent variable, together with the missingness pattern. We fit our model by maximizing $\log p_{\theta,\phi}(\xo, \zo)$ as described in Section \ref{sec:dgms_missing}.

We propose the following factorization of the joint distribution:
\begin{equation} \label{eq:moarm_factorization}
    p_{\theta, \phi}(\xo, \xm, \zo) = p_\theta(\xo)\, p_\theta(\xm \mid \xo)\, p_\phi(\zo \mid \xo, \xm).
\end{equation}
Note that this factorization is only relevant at inferece time, i.e., when $\xo$ and $\zo$ are available. At generation time, to ensure causality, we would instead assume fully observed data, i.e., $\x := \xo$ (equivalently, $\xm = \emptyset$), which recovers the standard OA-ARM formulation shown in Figure~\ref{fig:graph_oarm}. The factorization proposed in Equation (\ref{eq:moarm_factorization}) yields
\begin{equation}
    p_{\theta,\phi}(\xo, \zo)
    = p_\theta(\xo)
      \int p_\theta(\xm \mid \xo)\,
           p_\phi(\zo \mid \xo, \xm)\, d\xm.
\end{equation}
The inference structure assumes that the missingness pattern may depend on both $\xo$ and $\xm$, enabling MNAR mechanisms. In the special case of MCAR data, the missingness mechanism is independent of the data, and the term $p_\phi(\zo \mid \xo, \xm)$ can therefore be omitted during training.
Using Jensen’s inequality, we obtain the variational lower bound:
\begin{equation}
    \log p_{\theta,\phi}(\xo, \zo)
   	\ge 
        \log p_\theta(\xo)
      + 
      \mathbb{E}_{p_\theta(\xm \mid \xo)}
      \big[
        \log p_\phi(\zo \mid \xo, \xm)
      \big]
    \label{eq:moarm_elbo_labeled}
\end{equation}


To operationalize the factorization in  \eqref{eq:moarm_factorization}, we use two OA-ARM generative processes: one for $\xo$, and one for $\xm$ conditioned on $\xo$. Let $L = \Lo + \Lm$ denote the sizes of the observed and missing subsets, and partition the unmasking variables into $(\zo, \zm)$, where $\zo = \z_{\leq \Lo}$ and $\zm=\z_{> \Lo}$. The first term, i.e. the log-likelihood of the observed part, is obtained via the OA-ARM Markovian formulation:
\begin{equation}
    p_\theta(\xo, \zo) = \prod_{i=1}^{\Lo} p(z_i \mid \z_{<i}) \, p_\theta(\x_{z_i} \mid \x_{\z_{<i}}),
\end{equation}
We estimate this expression with the OA-ARM objective presented in \eqref{eq:oarm_obj}.

\paragraph{Missingness process}

The distribution of the missing part $\x_m$ is derived as conditional OA-ARM on the observed part:
\begin{equation} \label{eq:moarm_missing}
    p_\theta(\xm, \zm \mid \xo, \zo) = 
    \prod_{i=\Lo + 1}^{L} p(z_i \mid \z_{<i}) \, p_\theta(\x_{z_i} \mid \x_{\z_{<i}}).
\end{equation}
We further parameterize the missingness mechanism as an element-wise Bernoulli model on the binary indicator representation of $\zo$. Although $\zo$ is defined as the set of observed indices, it is in one-to-one correspondence with its binary indicator vector in $\{0,1\}^L$. The Bernoulli factorization below is written for this binary representation:
\begin{equation}\label{eq:moarm_zo_given_x}
    p_\phi(\zo \mid \xo,\xm)
    = \prod_{j=1}^{L}
        \mathrm{Bern}\!\left(
            \mathbbm{1}\{j \in \zo\} ; \pi_{\phi,j}(\xo,\xm)
        \right),
\end{equation}
where $\mathbbm{1}\{j \in \zo\}$ indicates whether element $j$ is present in $\zo$, and each $\pi_{\phi,j}$ represents the probability that element $j$ is observed. In this work, we adopt a neural-network parameterization of $\pi_{\phi,j}$ for maximal flexibility, enabling \model{} to capture general general missing patterns while keeping the implementation simple and scalable.

Using \eqref{eq:moarm_missing} and \eqref{eq:moarm_zo_given_x}, the second term in the ELBO is estimated via Monte Carlo using samples from the conditional distribution $p_\theta(\xm \mid \xo)$. In practice, we found it important to anneal the influence of the missingness likelihood on the generative model early in training: the missingness network is trained throughout, but gradients from $\log p_\phi(\zo \mid \xo,\xm)$ into $\theta$ are multiplied by a factor $\alpha_t$ that increases from $0$ to $1$. This prevents early collapse toward completions that explain the mask while yielding poor imputations. The resulting objective is optimized with a minibatch-parallel training procedure; we provide the full training algorithm in Appendix~\ref{sec:app_training_algorithm}.

\subsection{Imputation with \model{}} \label{sec:imputation}

Leveraging the conditional generation mechanisms described in Section~\ref{sec:moarm_prob_model}, \model{} performs imputation by marginalizing over possible unmasking trajectories of the missing variables. In the MCAR setting, the expectation of the missing part $\xm$ given the observed part $\xo$ can be written as
\begin{equation}
    \E\left[\xm \mid \xo, \zo \right]
    =
    \E_{p(\xm, \zm \mid \xo, \zo)}
    \left[p_\theta(\xm \mid \xo, \zo, \zm) \right].
\end{equation}
This expectation can be efficiently approximated using Monte Carlo sampling:
\begin{equation}
    \E\left[\xm \mid \xo, \zo \right]
    \approx
    \frac{1}{K} \sum_{k=1}^K
    \E\left[ \xm \mid \xo, \zo, \zm^{(k)}\right],
\end{equation}
where each completion $\xm^{(k)} \sim p_\theta(\xm \mid \xo, \zm^{(k)})$ is obtained by drawing an unmasking trajectory $\zm^{(k)} \sim p(\zm \mid \zo)$ and following the autoregressive recursion in Equation~\ref{eq:moarm_missing}.

For MNAR data, however, the observed mask $\zo$ itself carries information about the missing values. When \model{} is trained with an explicit missingness model $p_\phi(\zo \mid \x)$, we use this model at imputation time to reweight the Monte Carlo completions. Given completed samples
$
    \hat{\x}^{(k)} = (\xo, \xm^{(k)}),
$
we assign each completion an importance weight proportional to the likelihood of the observed mask under the missingness model:
\begin{equation}
    \tilde{w}_k
    =
    p_\phi\left(\zo \mid \hat{\x}^{(k)}\right),
    \qquad
    w_k
    =
    \frac{\tilde{w}_k}{\sum_{\ell=1}^K \tilde{w}_\ell}.
\end{equation}
The MNAR-aware imputation estimate is then the self-normalized importance weighted average
\begin{equation}
    \E\left[\xm \mid \xo, \zo \right]
    \approx
    \sum_{k=1}^K w_k \xm^{(k)}.
\end{equation}
Thus, completions under which the actually observed missingness pattern is more likely receive larger weight. In the absence of a learned missingness model, or under ignorable missingness, this reduces to the uniform Monte Carlo average above.

\paragraph{Parallelization} Although different observations may contain different numbers of missing variables, these trajectories can still be parallelized by batching together trajectories with the same number of currently revealed elements. We describe this bucketed parallelization strategy in Appendix~\ref{sec:app_parallel_imputation}.

\subsection{Efficient Data Acquisition with \model{}} \label{sec:saia}

Because \model{} amortizes all conditional distributions of the form
$p_\theta(\x_A | \x_B)$, it is naturally suited for settings where
only part of the input is observed. In \emph{Active Feature Acquisition} (AFA) \citep{melville2004active, saar2009active, huang2018active}, the goal is to sequentially reveal unobserved variables of $\x$ under a budget constraint, selecting at each step the variable that most improves prediction of a target $\x_\Phi$. In our tabular data experiments, we incorporate the target $\x_\Phi$ to the set of features in $\x$ when training the models.

At step $i$, let $\z_{<i}$ denote the set of already
revealed indices, and let $z_i \in \z_{\ge i}$ be a candidate next
variable to acquire. Following work on Sequential Active Information
Acquisition (SAIA) \citep{ma2019eddi, ma2020vaem, peis2022missing}, we
evaluate the usefulness of acquiring $z_i$ through the conditional
mutual information:
\begin{equation} \label{eq:mi_reward}
    R(i, z_i)
    =
    \mathcal{I}\!\left(
        \x_\Phi \,;\, \x_{z_i}
        |
        \x_{\z_{<i}} \right)
\end{equation}
which measures how much observing $\x_{z_i}$ would reduce uncertainty about the target given the current observations. Although estimating the mutual information in Eq. \ref{eq:mi_reward} is generally intractable, we adopt sampling-based estimators from prior work \cite{peis2022missing}, which have been shown to be effective in active feature acquisition with deep generative models. Details are provided in Appendix \ref{sec:app_mi_estimation}.


\section{Experiments}\label{sec:exps}

\begin{figure*}[t]
    \centering
    \newlength{\imgwidth}
    \setlength{\imgwidth}{0.75cm}
    \newlength{\sampleswidth}
    \setlength{\sampleswidth}{\dimexpr 5\imgwidth\relax}
    \newlength{\plotwidth}
    \setlength{\plotwidth}{\dimexpr 5.8\imgwidth\relax}
    \begin{subfigure}[t]{\imgwidth}
        \centering
        \includegraphics[width=\linewidth]{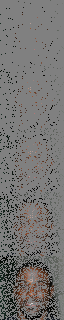}
        \caption*{$\xo$}
    \end{subfigure}\hfill
    \begin{subfigure}[t]{\sampleswidth}
        \centering
        \includegraphics[width=\linewidth]{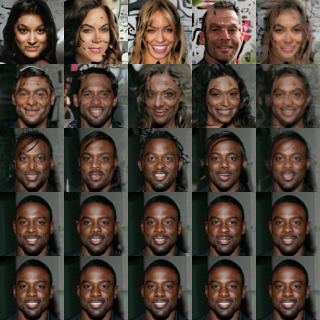}
        \caption*{$\xm\mid\xo$ (DDPM/RePaint)}
    \end{subfigure}\hfill
    \begin{subfigure}[t]{\sampleswidth}
        \centering
        \includegraphics[width=\linewidth]{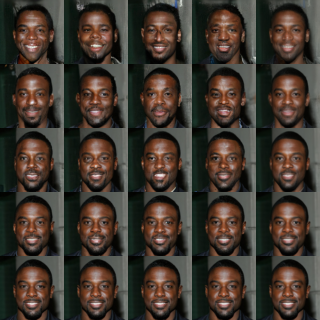}
        \caption*{$\xm\mid\xo$ (\model)}
    \end{subfigure} \hfill
    \begin{subfigure}[t]{\imgwidth}
        \centering
        \includegraphics[width=\linewidth]{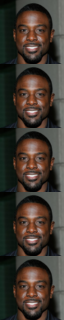}
        \caption*{$\x$}
    \end{subfigure}\hfill
    \begin{subfigure}[t]{\plotwidth}
        \centering
        \includegraphics[width=\linewidth]{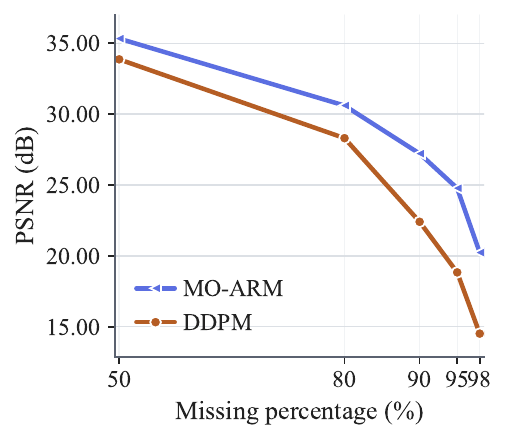}
        \caption*{PSNR}
    \end{subfigure}
    \caption{Image inpainting comparison. The leftmost panel shows incomplete images at different rates. The second and third panels show four columns with conditional samples produced by RePaint and \model{}, respectively, with their averages in the fifth columns. The fourth panel shows the ground truth image. The rightmost panel reports average PSNR as a function of the missing percentage.}
    \label{fig:image_inpainting_comparison}
\end{figure*}

\subsection{Experimental setup} \label{sec:exp_setup}

\paragraph{Datasets.}
We evaluate \model{} on both tabular imputation benchmarks and image inpainting.
For tabular data, we use seven real-world benchmarks from the UCI repository \cite{dheeru2017uci}: Adult, Bean, California, Default, Gesture, Letter and Magic.
The datasets span a variety of scales and feature types, including purely continuous variables (Gesture, Magic)
and mixed continuous-categorical features (Adult, Bean, California, Default, Letter).
All models are trained on the full joint distribution $p(\bm{x}, \bm{x}_\Phi)$, treating the target $\bm{x}_\Phi$ as an ordinary feature. For image inpainting, we use the CelebA-HQ dataset \cite{liu2015deep} resized to $64 \times 64$ RGB resolution.
Images are normalized to $[-1,1]$ channel-wise, and models are trained on the full image distribution without class labels.

\paragraph{Missing mechanisms and rates.}
For tabular data, we evaluate two missingness mechanisms that represent two ends of the standard missing-data spectrum.
Under MCAR, each entry is independently masked with probability $p$.
Under MNAR, we employ self-masking via a logistic function of each variable's own value, so that missingness depends directly on the unobserved signal.
We consider three training missing rates, $\{10\%, 30\%, 50\%\}$, and evaluate imputation performance at a fixed test missing rate of $50\%$.
For each combination of dataset, mechanism, and rate, we generate 5 independent missingness masks, train a separate model per mask, and report mean~$\pm$~standard deviation across the 5 splits.

\paragraph{Baselines.}
For the imputation benchmarks, we compare \model{} with four closely related generative baselines:
(i) MIWAE~\cite{mattei2019miwae}, a VAE for incomplete data trained with an importance-weighted ELBO;
(ii) not-MIWAE~\cite{ipsennot}, a variant of MIWAE to model MNAR mechanisms;
(iii) VAEAC~\cite{ivanov2018variational}, a VAE with an arbitrarily conditioned prior;
and (iv) DiffPuter~\cite{zhang2025diffputer}, a robust EM-based diffusion imputer.
For completeness, we also include classical deterministic baselines such as MissForest~\cite{stekhoven2012missforest} and MICE~\cite{van2011mice}, as well as the more recent HyperImpute method~\cite{jarrett2022hyperimpute}.
For the SAIA setting, we restrict comparisons to generative models that can provide samples from the required conditional distributions.

\paragraph{Training.}
Tabular models are trained on a single NVIDIA V100 GPU, while image models are trained on NVIDIA H100 GPUs. 
Full architecture specifications and hyperparameters are provided in Appendix~\ref{sec:app_exp_details}. Our code is accessible at \href{https://anonymous.4open.science/r/MOARM_pre-B310/}{this link}.

\subsection{Image Inpainting} \label{sec:exp_inpainting}
We include a qualitative image inpainting experiment as an illustrative sanity check outside the main tabular setting.
We train \model{} on fully observed CelebA-HQ images, corresponding to an OA-ARM instance with no missing entries during training.
Since exact OA-ARM sampling would require one network evaluation per missing pixel, direct sampling would is expensive at image dimensionalities.
We therefore use the exponential blocked-unmasking schedule described in Appendix~\ref{sec:app_sampling_schedule}, which reveals small blocks early and larger blocks later; all \model{} inpaintings in this section are generated using only 64 sampling steps.
At test time, we condition on sparse observed pixels, varying the missing percentage in $\{50\%, 80\%, 90\%, 95\%, 98\%\}$, and compare the resulting conditional samples with RePaint~\cite{lugmayr2022repaint} applied to a standard unconditional DDPM~\cite{ho2020denoising}, providing a robust diffusion-based reference for the same fully observed training setting.

Figure~\ref{fig:image_inpainting_comparison} shows that both methods generate plausible completions from randomly observed pixel contexts.
However, \model{} more consistently preserves the identity and coarse structure of the ground-truth image across posterior samples, whereas RePaint exhibits greater variation in facial attributes and pose.
The PSNR curve supports this qualitative trend: \model{} achieves higher reconstruction accuracy than the DDPM/RePaint reference at every evaluated MCAR missing rate, and the difference increases as the observed context becomes more sparse.
Together, these results suggest that the OA-ARM formulation can produce accurate conditional predictions under random pixel missingness while still maintaining meaningful sample diversity.

\subsection{Tabular Imputation} \label{sec:exp_imputation}

\begin{figure}[t]
    \centering
    \includegraphics[width=\linewidth]{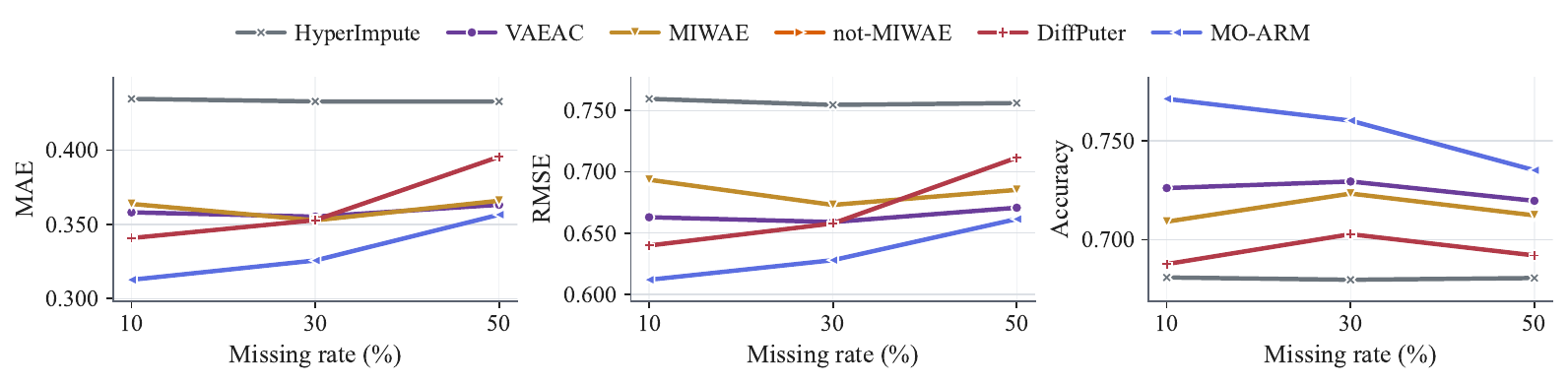}
    \includegraphics[width=\linewidth]{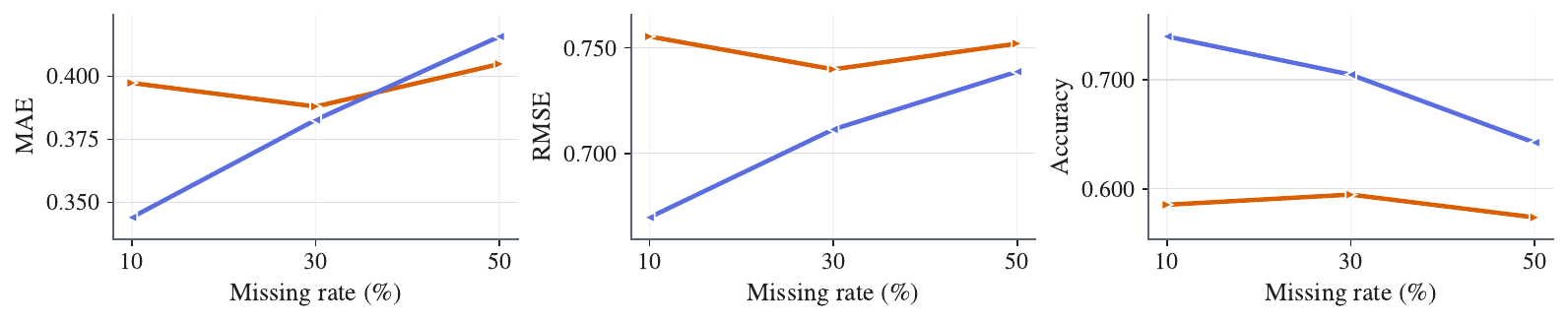}
    \caption{
    Average test imputation performance at a 50\% missing rate across the nine UCI tabular benchmarks. The horizontal axis denotes the training missing rate. For the accuracy metric, results are reported on the subset of four datasets containing categorical features. The top row corresponds to MCAR missingness, while the bottom row corresponds to MNAR missingness.}    \label{fig:metrics_average}
\end{figure}

We next evaluate whether \model{} can use its conditional generative structure to impute missing entries in heterogeneous tabular data. Imputation is performed using $K=100$ conditional samples. For \model,  we use the estimator described in section \ref{sec:imputation}. We report RMSE and MAE for continuous features, and accuracy for categorical features after decoding the binary representation back to category indices. All metrics are computed exclusively on the held-out missing entries.

It is worth noting, however, that these methods differ substantially in computational cost. In particular, DiffPuter relies on an EM-style training, and requires diffusion-based inference for each imputation, making it considerably more expensive than the amortized conditional sampling used by \model{}.

Figure~\ref{fig:metrics_average} summarizes the average imputation performance across datasets as a function of the training missing rate, focusing on the strongest baselines for readability; the full per-dataset results and complete baseline comparison are reported in Appendix~\ref{sec_app:exp}, Figures~\ref{fig:app_tab_rmse}--\ref{fig:app_tab_MNAR_disc}. Under MCAR missingness, \model{} performs best at all missing rates, consistently improving over established and recent imputers. This behavior aligns with the connection developed in Section~\ref{sec:moarm_prob_model}: order-agnostic training naturally exposes the model to randomly observed subsets, making it especially well matched to MCAR imputation. Under MNAR missingness, \model{} outperforms not-MIWAE, the only baseline in our benchmark explicitly designed to model MNAR data.

\begin{figure*}[t]
    \centering
    \begin{subfigure}{\linewidth}
    \centering
    \includegraphics[width=0.45\linewidth]{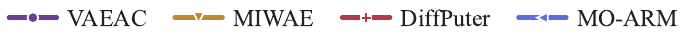}
    \end{subfigure}\hfill
    
    \def\imgwidth{0.24\textwidth}
    
    \begin{subfigure}{\imgwidth}
        \centering
        \includegraphics[width=\linewidth]{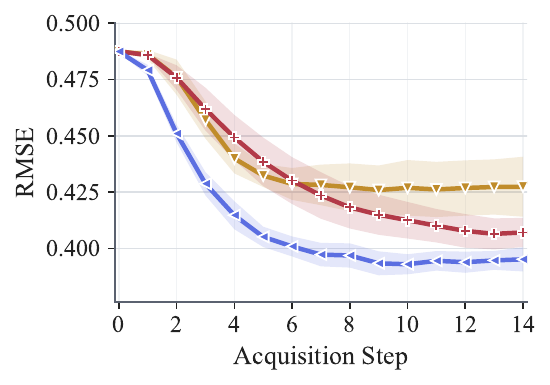}
        \caption{Adult}
    \end{subfigure}\hfill
    \begin{subfigure}{\imgwidth}
        \centering
        \includegraphics[width=\linewidth]{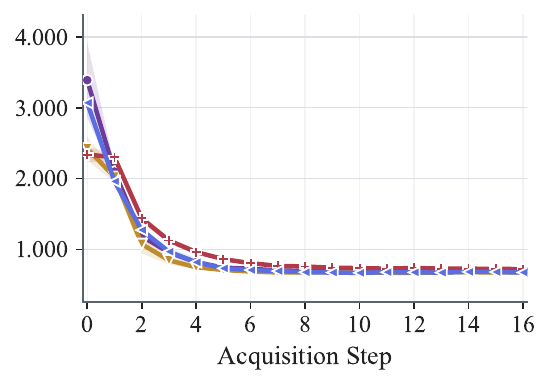}
        \caption{Bean}
    \end{subfigure}\hfill
    \begin{subfigure}{\imgwidth}
        \centering
        \includegraphics[width=\linewidth]{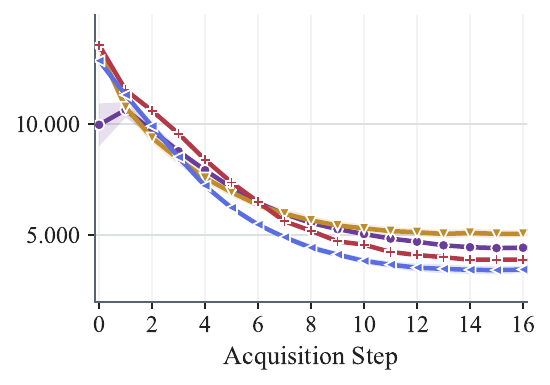}
        \caption{Letter}
    \end{subfigure}    
    \begin{subfigure}{\imgwidth}
        \centering
        \includegraphics[width=\linewidth]{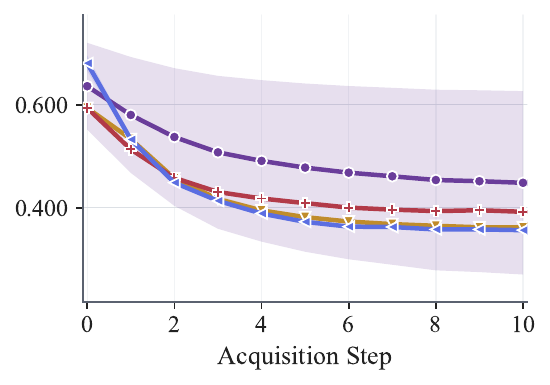}
        \caption{Magic}
    \end{subfigure}\hfill
        \caption{
            Sequential active information acquisition with \model{} and baselines on UCI datasets.
            }
		\label{fig:saia}
\end{figure*}

\subsection{Sequential Active Information Acquisition}

We next test whether the models trained for imputation can also support sequential active information acquisition. We reuse the MCAR models trained with a $30\%$ missing rate. Starting from an empty observed set, each method sequentially acquires one feature while the target $\x_\Phi$ remains hidden. As in Section~\ref{sec:saia}, features are ranked by estimated conditional mutual information with the target; following \cite{peis2022missing}, we approximate this quantity using $100$ conditional samples and $5$ discretization bins. After each acquisition, we update the predictive distribution of $\x_\Phi$ and report prediction error.

Figure~\ref{fig:saia} shows that error decreases steadily as more variables are acquired for all methods, with \model{} consistently matching or outperforming the baselines. In particular, \model{} consistently outperforms DiffPuter throughout the acquisition trajectory, while MIWAE exhibits less consistent behavior: it can be competitive at the first few acquisition steps, but is generally overtaken at later steps as more informative variables are revealed. These results show that the same amortized conditional model used for imputation can also guide effective feature acquisition, supporting a unified treatment of imputation and sequential decision making under partial observability.

\section{Limitations}
\label{sec:limitations}

Our approach has several limitations. Exact OA-ARM generation remains expensive in high-dimensional settings because it requires many autoregressive steps, although blocked unmasking partly mitigates this cost (Section~\ref{sec:app_sampling_schedule}). Imputation quality also deteriorates as observed context becomes very sparse, making \model{} most effective at low-to-moderate missing rates in our experiments. Finally, while the missingness-aware formulation extends to MNAR data, this regime remains more challenging than MCAR and deserves further study.

\section{Conclusion}

We introduced \model{}, a missingness-aware order-agnostic autoregressive framework for reasoning under incomplete observations. By connecting order-agnostic training to MCAR imputation, extending OA-ARMs to learn directly from incomplete data under general missingness mechanisms, and reusing the same conditional model for active information acquisition, our approach provides a unified treatment of imputation and sequential feature acquisition. Empirically, \model{} achieves superior performance than the alternatives on heterogeneous tabular benchmarks. More broadly, our results suggest that order-agnostic autoregressive models offer a practical and principled alternative to latent-variable and diffusion-based approaches for missing-data problems: they retain tractable likelihoods, support arbitrary conditioning sets, and can be queried repeatedly as observations are revealed. We hope this perspective encourages further work on autoregressive models as general-purpose tools for decision making under partial observability.


\bibliography{main}
\bibliographystyle{plain}

\newpage
\appendix

\section{Further theoretical analysis}

\subsection{Mutual Information estimation} \label{sec:app_mi_estimation}

Our approach follows a widely used and well-understood method for estimating 
$\mathcal{I}\!\left(
        \x_\Phi \,;\, \x_{z_i}
        | \x_{\z_{<i}} \right)$, 
detailed in \citep{kraskov2004estimating}. Specifically, we discretize the continuous variables $\x_{\Phi}$ (target variable) and $\x\left[\Delta_i\right]$ (candidate feature) into finite uniform bins and estimate MI via empirical co-occurrence frequencies:

\begin{equation}
    \mathcal{I}\!\left(
        \x_\Phi \,;\, \x_{z_i}
        |
        \x_{\z_{<i}} \right)
    \approx \sum_{u\in B_\Phi}\sum_{v\in B_{z_i}}  \pi(u, v) \log \frac{\pi(u, v)}{\pi_{x_{\Phi}}(u) \, \pi_{x_{z_i} }(v)},
\end{equation}
where $B_\Phi$ and $B_{z_i}$ denote the discretization bins. The binned probabilities are defined as 
$$
\pi_{x}(b) =  \int_b p_{x}(x) dx, 
$$
for the interval defined by bin $b$, and can be estimated from $N$ samples via:
\begin{equation}
    \pi_{x_{\Phi}}(u) \approx \frac{n_{x_{\Phi}}(u)} {N}, \quad
\pi_{x_{z_i}}(v) \approx \frac{n_{x_{z_i}}(v)} {N}, \quad
\pi(u,v) \approx \frac{n(u, v)} {N},
\end{equation}
where $n_{x_{\Phi}}(u)$ and $n_{x_{z_i}}(v)$ denote the number of samples assigned to bins $u$ and $v$, respectively, and $n(u,v)$ denotes the number of samples jointly assigned to bin $(u,v)$.

This method has been shown to converge to the true MI as $N \to \infty$ under standard assumptions, and, while simple, it has previously been validated in the context of Active Feature Acquisition (AFA) in \citep{peis2022missing}.

\subsection{Accelerated Sampling in OA-ARMs}
\label{sec:app_sampling_schedule}

Standard OA-ARM sampling reveals one variable per autoregressive step,
requiring $L$ evaluations of the conditional model. While exact, this procedure is slow for high-dimensional data.
At the same time, OA-ARMs exhibit a striking empirical property:
global structure emerges extremely early in the autoregressive
trajectory. As shown in Figure~\ref{fig:image_inpainting_comparison}, even when only a small fraction of variables has been revealed, the model already
reconstructs coherent semantic structure (digit identity, object
silhouette, facial layout), and subsequent steps primarily refine local
details. This coarse-to-fine behaviour arises naturally from training
over random partial contexts, many of which expose highly informative
subsets at early steps.

Motivated by this observation, we propose a blocked unmasking
procedure that reveals only a few variables at early steps and
increasingly larger blocks later. Rather than unmasking a single
variable $z_i$ at every iteration, we unmask a block $B_t$ of $k_t$
indices at step $t$, where the block sizes follow an exponential
schedule. Formally, we choose $k_t = \lfloor \alpha^t \rfloor$ for a
scalar $\alpha > 1$ determined by the constraint
$\sum_{t=1}^{S} k_t = L$, where $S \ll L$ is the desired number of
sampling steps. Solving for $\alpha$ yields block sizes of the form
$k_t \approx L^{t/S}$, which produce small updates early and rapidly
growing updates later. The resulting cumulative number of revealed
variables then satisfies $s_t \approx L^{t/S}$, enabling the model to
complete generation in exactly $S$ steps.

Given a block $B_t = \{ z_{s_{t-1}+1}, \dots, z_{s_t} \}$, we define the
blocked likelihood contribution as
\begin{equation}
    \tilde{p}_\theta(\x | \z)
    := \prod_{t=1}^{S}
        \prod_{j \in B_t}
        p_\theta\bigl(\x_j | \x_{\z_{\leq s_{t-1}}}\bigr),
\end{equation}
which preserves the autoregressive semantics of OA-ARMs while reducing
the number of model evaluations by a factor of $L/S$. Importantly, early
blocks remain small, preserving the model’s ability to infer global
structure from limited context, while late blocks are large, accelerating
the refinement phase where the model primarily resolves local details.

In practice, our proposed scheme produces $32$--$64\times$
speed-ups in wall-clock sampling time with negligible degradation in perceptual quality. This establishes exponential unmasking as a principled and highly effective acceleration mechanism for OA-ARMs, fully compatible with the probabilistic formulation presented in Section~\ref{sec:moarm_prob_model}.

\subsection{Parallel Imputation with \model{}}
\label{sec:app_parallel_imputation}

The accelerated sampling scheme above assumes that all samples begin from the
same initial state, namely the empty context. Imputation is more subtle:
different incomplete observations may contain different numbers of observed
variables. If two samples have observed-index sets $\zo^{(a)}$ and
$\zo^{(b)}$ with $|\zo^{(a)}| \neq |\zo^{(b)}|$, then their conditional
generation processes start at different points along the autoregressive
trajectory. A naive implementation would therefore run one chain per sample,
or pad all chains to the longest trajectory and mask out completed samples.
Both choices are inefficient, either because they prevent effective batching
or because they spend computation on trajectories that have already finished.

\model{} avoids this inefficiency by grouping imputation trajectories
according to their current number of revealed variables. Let $n$ index data
points and $k$ index Monte Carlo imputations. At an intermediate stage of
imputation, each trajectory has a revealed set
$\z_{\leq s}^{(n,k)}$ containing the originally observed variables together
with any variables imputed so far. We define the cardinality bucket
\begin{equation}
    \mathcal{C}_s
    =
    \left\{
        (n,k) :
        \left| \z_{\leq s}^{(n,k)} \right| = s
    \right\}.
\end{equation}
All trajectories in $\mathcal{C}_s$ are at the same autoregressive state in
terms of reveal count, even though the identities of the revealed variables
may differ. Hence, the corresponding conditionals
\begin{equation}
    p_\theta
    \left(
        \x_j^{(n,k)}
        \mid
        \x_{\z_{\leq s}^{(n,k)}}
    \right),
    \qquad
    j \notin \z_{\leq s}^{(n,k)},
    \qquad
    (n,k)\in\mathcal{C}_s,
\end{equation}
can be evaluated in a single batched network call. The masks
$\z_{\leq s}^{(n,k)}$ may vary across the batch; only their cardinality must
match.

This bucketed implementation is directly compatible with the accelerated
blocked unmasking procedure from Section~\ref{sec:app_sampling_schedule}.
For a bucket $\mathcal{C}_s$, we choose a block size $k_t$ from the sampling
schedule and reveal, for each trajectory, a block
\begin{equation}
    B_t^{(n,k)}
    \subseteq
    \{1,\dots,L\} \setminus \z_{\leq s}^{(n,k)},
    \qquad
    |B_t^{(n,k)}| = k_t.
\end{equation}
The variables in this block are predicted conditionally on the current
revealed set,
\begin{equation}
    \prod_{j \in B_t^{(n,k)}}
    p_\theta
    \left(
        \x_j^{(n,k)}
        \mid
        \x_{\z_{\leq s}^{(n,k)}}
    \right),
\end{equation}
and the revealed set is updated as
\begin{equation}
    \z_{\leq s+k_t}^{(n,k)}
    =
    \z_{\leq s}^{(n,k)}
    \cup
    B_t^{(n,k)}.
\end{equation}
The trajectory is then moved to the bucket corresponding to its new revealed
cardinality. Trajectories with fewer missing variables terminate earlier,
while trajectories with more missing variables continue through later
buckets. Thus, no forward passes are wasted on completed imputations.

Monte Carlo imputation is parallelized by the same mechanism. Each incomplete
observation is replicated $K$ times along the batch dimension, and each
replica follows an independent unmasking trajectory. These replicas are then
bucketed together with all other trajectories according to their current
revealed cardinality. Consequently, heterogeneous missingness across data
points and uncertainty over imputation trajectories are both handled through
ordinary batching.

In practice, this means that imputation with \model{} is not implemented as
one sequential autoregressive process per sample. Instead, we repeatedly
gather all active trajectories with the same number of revealed variables,
evaluate them together, reveal the next scheduled block, and reassign them to
the appropriate cardinality bucket. This preserves the order-agnostic
conditioning structure of OA-ARMs while enabling efficient GPU utilization
even when samples contain different numbers of missing elements.

\subsection{Training algorithm for \model{}}
\label{sec:app_training_algorithm}

Algorithm~\ref{alg:moarm_training} summarizes training of \model{} on incomplete data. Each minibatch contains partially observed examples $\xo$ together with their observed-index sets $\zo$. The observed-data term is estimated with the single-step OA-ARM estimator from Eq.~\eqref{eq:oarm_obj}, applied in parallel across the minibatch. When the missingness mechanism is informative, we additionally sample Monte Carlo completions of the missing variables from the conditional OA-ARM and train the missingness model $p_\phi(\zo \mid \xo,\xm)$. For MCAR data, missingness is independent of the values and this second term is omitted.

\begin{algorithm}[t]
\caption{Minibatch training of \model{} on incomplete data}
\label{alg:moarm_training}
\begin{algorithmic}[1]
\Require Dataset $\mathcal{D}=\{(\xo^{(n)},\zo^{(n)})\}_{n=1}^N$, number of Monte Carlo samples $K$, learning rate $\eta$
\Require OA-ARM parameters $\theta$, optional missingness-model parameters $\phi$, annealing schedule $\alpha_t \in [0,1]$
\Repeat
    \State Sample a minibatch $\mathcal{B}=\{(\xo^{(b)},\zo^{(b)})\}_{b=1}^B$
    \State Let $\Lo^{(b)} = |\zo^{(b)}|$ for each minibatch element
    \State Sample $i^{(b)} \sim \mathrm{Unif}\{1,\dots,\Lo^{(b)}\}$ independently for all $b$
    \State Sample context sets $\z_{<i}^{(b)} \subset \zo^{(b)}$ uniformly with $|\z_{<i}^{(b)}|=i^{(b)}-1$
    \State Compute the observed-data OA-ARM estimator in parallel:
    \[
    \widehat{\mathcal{L}}_{\mathrm{obs}}
    =
    \frac{1}{B}
    \sum_{b=1}^{B}
    \frac{\Lo^{(b)}}{\Lo^{(b)}-i^{(b)}+1}
    \sum_{j\in \zo^{(b)}\setminus \z_{<i}^{(b)}}
    \log p_\theta
    \left(
        x_j^{(b)}
        \mid
        \x_{\z_{<i}^{(b)}}^{(b)}
    \right).
    \]
    \If{the missingness mechanism is modeled}
        \State Replicate the minibatch $K$ times along the sample dimension
        \State Sample $K$ conditional completions in parallel:
        \[
        (\xm)^{(b,k)}
        \sim
        p_\theta
        \left(
            \xm
            \mid
            \xo^{(b)}, \zo^{(b)}
        \right).
        \]
        \State Form completions
        \[
        \hat{\x}^{(b,k)} = (\xo^{(b)},(\xm)^{(b,k)}).
        \]
        \State Form annealed completions
        \[
        \hat{\x}_{\alpha_t}^{(b,k)}
        =
        \mathrm{stopgrad}\!\left(\hat{\x}^{(b,k)}\right)
        +
        \alpha_t
        \left(
            \hat{\x}^{(b,k)}
            -
            \mathrm{stopgrad}\!\left(\hat{\x}^{(b,k)}\right)
        \right),
        \]
        \State Estimate the missingness-model term:
        \[
        \widehat{\mathcal{L}}_{\mathrm{miss}}
        =
        \frac{1}{BK}
        \sum_{b=1}^{B}
        \sum_{k=1}^{K}
        \log p_\phi
        \left(
            \zo^{(b)}
            \mid
            \hat{\x}_{\alpha_t}^{(b,k)}
        \right).
        \]
    \Else
        \State Set $\widehat{\mathcal{L}}_{\mathrm{miss}} \gets 0$
    \EndIf
    \State Form the minibatch objective:
    \[
    \widehat{\mathcal{L}}
    =
    \widehat{\mathcal{L}}_{\mathrm{obs}}
    +
    \widehat{\mathcal{L}}_{\mathrm{miss}}.
    \]
    \State Update parameters by gradient ascent:
    \[
    (\theta,\phi)
    \gets
    (\theta,\phi)
    +
    \eta
    \nabla_{\theta,\phi}
    \widehat{\mathcal{L}}.
    \]
\Until{convergence}
\end{algorithmic}
\end{algorithm}

\section{Experimental extension} \label{sec_app:exp}

\subsection{Tabular data details}\label{sec:app_tab_details}

We report the statistics of the considered tabular datasets in Table~\ref{tbl:stat-dataset}.

\begin{table}[ht] 
    \centering
    \caption{Statistics of datasets. \# Num and \# Cat denote the number of numerical and categorical columns, respectively, reported as feature plus target columns when target features are included.}
    \label{tbl:stat-dataset}
    \resizebox{\linewidth}{!}{
        \begin{tabular}{lccccc}
            \toprule[0.8pt]
            \textbf{Dataset}  &  \# Rows  & \# Num & \# Cat & \# Train (In-sample)  & \# Test (Out-of-Sample)  \\
            \midrule 
            \textbf{California} Housing  & $20,640$ & $9$ & $0+3$ & $14,303$ & $6,337$   \\
            \textbf{Letter} Recognition & $20,000$ & $16$ & $0+5$ & $14,000$ & $6,000$ \\
            \textbf{Gesture} Phase Segmentation & $9,522$ & $32$ & - & $6,665$ & $2,857$ \\
            \textbf{Magic} Gamma Telescope & $19,020$ & $10+1$ & - & $13,314$ & $5,706$  \\
            Dry \textbf{Bean} & $13,610$ & $16$ & $0+3$ & $9,527$ & $4,083$ \\
            \midrule
            \textbf{Adult} Income & $32,561$ & $6$ & $8+1$ & $22,792$ & $9,769$ \\
            \textbf{Default} of Credit Card Clients & $30,000$ & $14$ & $10+1$ & $21,000$ & $9,000$\\
            
            \bottomrule[1.0pt]
        \end{tabular}
    }
    \end{table}

In our implementation, we set \texttt{include\_target=True} for tabular experiments and model the joint distribution including the target variable.
Concretely, the target is appended as an additional feature:
if the target is categorical (or binary), it is binary-encoded and concatenated to the categorical block; if it is continuous, it is appended to the numerical block.
Therefore, the effective modeled dimensionality is
\begin{equation}
D_{\mathrm{model}} = D_{\mathrm{num}} + D_{\mathrm{cat,bin}} + D_{\mathrm{target,enc}}.
\end{equation}

Using the current dataset configurations, the resulting input dimensionalities are:
Adult ($35$), Bean ($19$), California ($12$), Default ($45$), Gesture ($32$), Letter ($21$), and Magic ($11$).

\subsection{Likelihood functions}

The parametric family used for the conditional likelihood depends on the data modality:
\begin{equation}
p_\theta(\x_{\z_i}\mid \x_{\z_{i-1}}).
\end{equation}
For heterogeneous tabular data (after preprocessing), we use Gaussian likelihoods over the encoded representation.
For binary images, we use Bernoulli likelihoods:
\begin{equation}
p_\theta(\x_{\z_i}\mid \x_{\z_{i-1}})
=
\mathcal{B}\!\left(f_\theta(\x_{\z_{i-1}})\right).
\end{equation}
For color images, following \cite{kingma2016improved}, we use a mixture of discretized logistic distributions, with component parameters predicted by the network:
\begin{equation}
\{\bm{\mu}_\ell,\bm{s}_\ell,\pi_\ell\}_{\ell=1}^{K}
=
f_\theta(\x_{\z_{i-1}}).
\end{equation}
For categorical tabular variables, each variable is represented by a fixed block of bits.
Predicted bit values are rounded and decoded back to category indices.

\subsection{General missing mechanisms}

We follow the missing data generation protocol and implementation from DiffPuter~\cite{zhang2025diffputer}.
For MCAR, each entry is independently masked with probability $p$.
For MNAR, we use the same ``logistic model with input masked by MCAR'' protocol as DiffPuter.
As in their setup, features are split into two groups: one group is used as input to a logistic model, which outputs missing probabilities for the other group.
After sampling missingness for this second group, we additionally apply MCAR masking to the input group.
Therefore, missingness in the second group depends on values that may themselves be masked, which induces MNAR missingness.

\subsection{Experimental details} \label{sec:app_exp_details}

\subsubsection{Tabular data preprocessing}
All tabular experiments use the same preprocessing and split protocol as described in Sec.~\ref{sec:exp_setup}: categorical variables are binary-encoded, continuous variables are standardized from observed training entries only, and models are trained/evaluated across 10 independent mask realizations per dataset--mechanism--rate combination.
Unless stated otherwise, optimization uses Adam-family optimizers, gradient clipping, and mixed-precision training.
For stochastic imputers, final predictions are obtained by Monte Carlo averaging with $K=100$ samples at test time.

We mainly follow the preprocessing of DiffPuter \cite{zhang2025diffputer}, categorical features are binary-encoded using $\lceil \log_2 C \rceil$ bits per category, where $C$ is the number of categories, yielding encoded dimensionalities ranging from 11 (Magic) to 52 (News).
Continuous features are z-score normalized using statistics computed exclusively from the observed entries of the training set.
Unlike DiffPuter, all models are trained on the full joint distribution $p(\bm{x}, \bm{x}_\Phi)$, treating the target $\bm{x}_\Phi$ as an ordinary feature, which enables principled uncertainty estimation over $\bm{x}_\Phi$.

\subsubsection{MO-ARM architecture and training.}
For tabular data, \model{} uses a time-conditioned MLP backbone with hidden widths $[512,1024,512]$ and time-embedding dimension 512 for the normalized autoregressive step $i/L$. 
The network receives the concatenation of data and context-mask features and predicts Gaussian parameters for all dimensions.
We train with AdamW (learning rate $4\times 10^{-4}$, $\beta=(0.9,0.999)$, weight decay $0$), batch size 1024, and up to 6000 epochs.
A ReduceLROnPlateau scheduler is used (factor $0.9$, patience $200$), with gradient clipping at 1.0.
An exponential moving average (EMA) of model parameters is maintained during training.
For MNAR-specific runs, we additionally optimize a missingness-prediction head with a BCE term jointly with the likelihood objective.

For CelebA-HQ, \model{} uses an image-specific U-Net backbone operating on $64\times64$ RGB images.
The input processor first concatenates the partially observed image with its binary context mask and maps the resulting tensor to 128 feature channels with a convolutional layer.
The U-Net is time-conditioned by the normalized autoregressive step $i/L$, uses 128 model channels, dropout 0.1, no attention layers, and outputs the parameters of a 30-component mixture of logistics likelihood for RGB pixels.
During training, images are fully observed, and random context masks are simulated internally so that the model learns conditional next-pixel distributions for arbitrary observed subsets.
We train with AdamW (learning rate $4\times 10^{-4}$, $\beta=(0.9,0.999)$, weight decay $10^{-2}$), batch size 64 across two NVIDIA H100 GPUs, and 32-bit precision.
Training is run for approximately $2\times10^6$ optimization steps using a cosine annealing learning-rate schedule, with gradient clipping at 1.0 and EMA parameters maintained throughout.

\subsubsection{Baseline details.}

\paragraph{VAEAC.}
VAEAC uses latent dimension $d_z=10$ and one Monte Carlo sample during training.
The encoder, prior network, and decoder are MLPs with hidden widths $[128,128,128]$, where both the encoder and prior network take concatenated data-mask inputs.
For continuous observations, we use a Gaussian likelihood with learned variance, implemented by predicting both mean and log-variance parameters.
We train with Adam (learning rate $10^{-3}$), batch size 256, for 3000 epochs, with gradient clipping at 1.0.

\paragraph{MIWAE.}
MIWAE uses latent dimension $d_z=10$ and $K=20$ importance samples during training.
Both the encoder and decoder are MLPs with hidden widths $[128,128,128]$, with the encoder taking concatenated data-mask inputs.
Following the original MIWAE formulation, we use Student-\(t\) distributions for both the variational posterior and the observation model.
We train with Adam (learning rate $10^{-3}$), batch size 256, for 3000 epochs, with gradient clipping at 1.0.

\paragraph{not-MIWAE.}
For not-MIWAE, we augment MIWAE with the same missingness prediction head as for MOARM. 


\paragraph{DiffPuter.}
DiffPuter is trained with an EM-style procedure.
We use 10 EM iterations and 10,000 training epochs per EM stage, with 20 samples in the E-step.
The diffusion backbone is an MLP-diffusion network with time-embedding dimension 1024.
Optimization uses Adam and a plateau scheduler (patience 50); the effective batch size is 4096.

\paragraph{Other tabular baselines.}
HyperImpute is run through its standard library interface with default robust settings used in prior work.
Classical deterministic imputers (e.g., MissForest, MICE, etc.) are included where applicable.

\paragraph{DDPM/RePaint.}
For the illustrative image inpainting experiment, we train an unconditional DDPM on fully observed CelebA-HQ images resized to $64\times64$.
The denoising network is a U-Net with 192 model channels, channel multipliers $[1,2,2,2]$, three residual blocks per resolution, dropout 0.1, and self-attention at resolutions $\{2,4,8\}$.
The diffusion process uses 1000 timesteps with a linear beta schedule from $10^{-4}$ to $2\times10^{-2}$, and the model predicts additive Gaussian noise.
We train with AdamW (learning rate $2\times10^{-4}$, weight decay $0$), batch size 64, 32-bit precision, gradient clipping at 1.0, and EMA parameters.
At test time, incomplete images are imputed using RePaint~\cite{lugmayr2022repaint} with jump length 10 and 5 resampling steps per jump.

\subsection{Per dataset imputation results}

\begin{figure*}[t]
    \centering
    \begin{subfigure}{\linewidth}
    \centering
    \includegraphics[width=\linewidth]{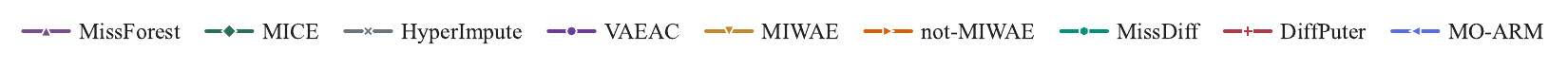}
    \end{subfigure}\hfill
    \def\imgwidth{0.24\textwidth}
    \begin{subfigure}{\imgwidth}
    \centering
    \includegraphics[width=\linewidth]{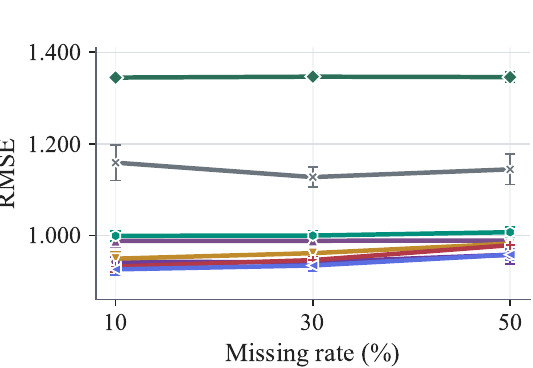}
    \caption{Adult}
    \end{subfigure}\hfill
    \begin{subfigure}{\imgwidth}
        \centering
    \includegraphics[width=\linewidth]{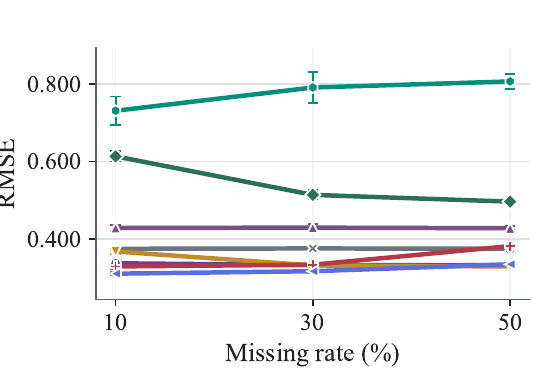}
        \caption{Bean}
    \end{subfigure}\hfill
    \begin{subfigure}{\imgwidth}
        \centering
        \includegraphics[width=\linewidth]{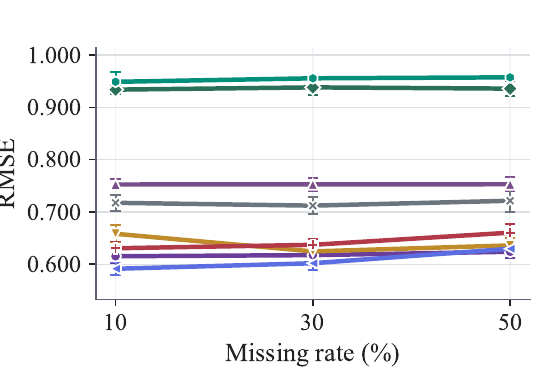}
        \caption{California}
    \end{subfigure}
    \begin{subfigure}{\imgwidth}
        \centering
        \includegraphics[width=\linewidth]{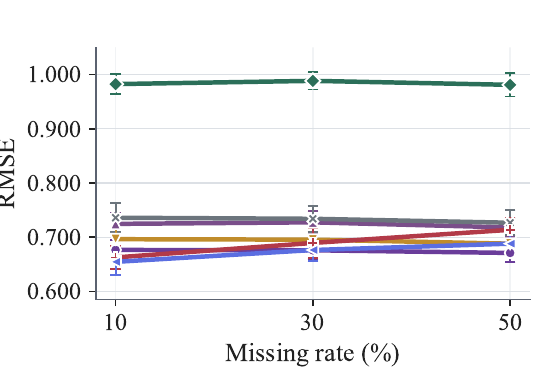}
        \caption{Default}
    \end{subfigure}\hfill
    \begin{subfigure}{\imgwidth}
        \centering
        \includegraphics[width=\linewidth]{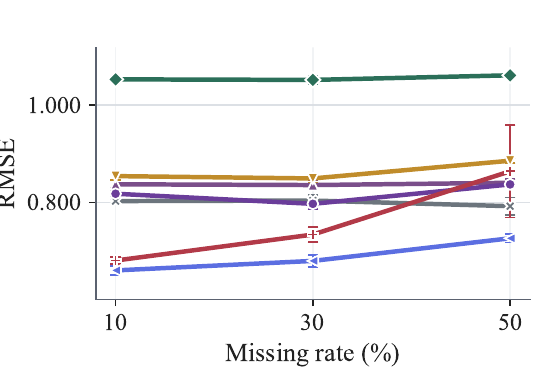}
        \caption{Gesture}
    \end{subfigure}
    \begin{subfigure}{\imgwidth}
        \centering
        \includegraphics[width=\linewidth]{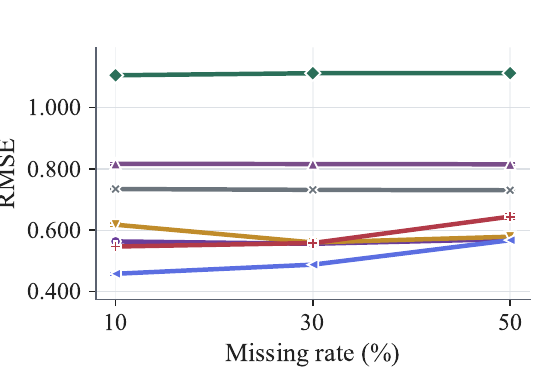}
        \caption{Letter}
    \end{subfigure}
    \begin{subfigure}{\imgwidth}
        \centering
        \includegraphics[width=\linewidth]{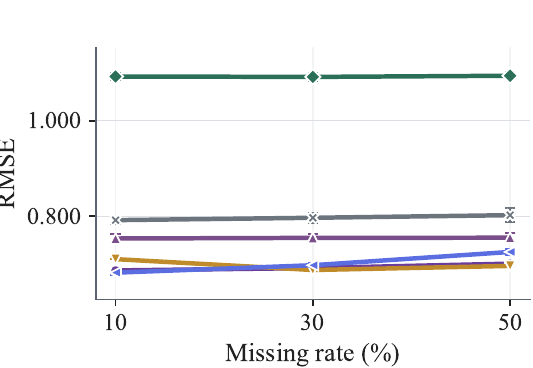}
        \caption{Magic}
    \end{subfigure}
    \caption{Per-dataset average test imputation RMSE of MCAR missing data at a 50\% missing rate across five missing masks. The horizontal axis indicates the training missing rate.}

    \label{fig:app_tab_rmse}
\end{figure*}

\begin{figure*}[t]
    \centering
    \begin{subfigure}{\linewidth}
    \centering
    \includegraphics[width=\linewidth]{figs/legend.pdf}
    \end{subfigure}\hfill
    \def\imgwidth{0.24\textwidth}
    \begin{subfigure}{\imgwidth}
    \centering
    \includegraphics[width=\linewidth]{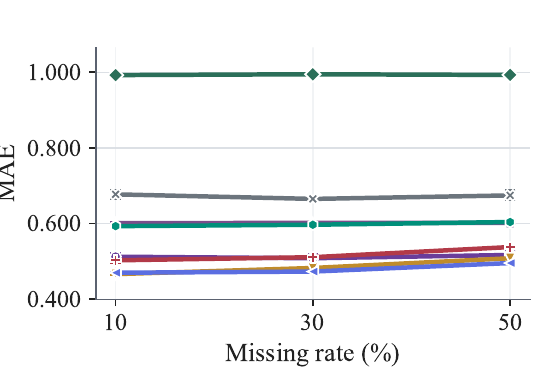}
    \caption{Adult}
    \end{subfigure}\hfill
    \begin{subfigure}{\imgwidth}
        \centering
    \includegraphics[width=\linewidth]{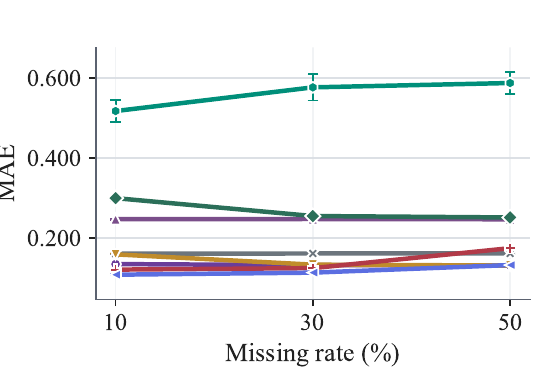}
        \caption{Bean}
    \end{subfigure}\hfill
    \begin{subfigure}{\imgwidth}
        \centering
        \includegraphics[width=\linewidth]{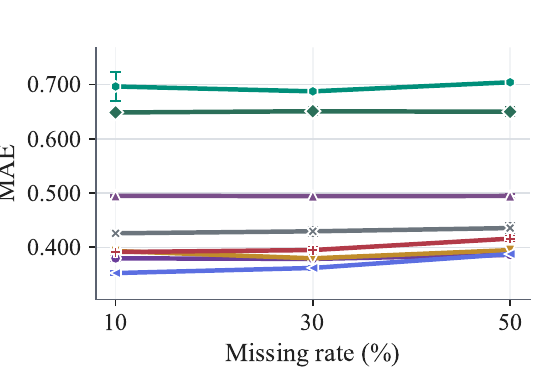}
        \caption{California}
    \end{subfigure}
    \begin{subfigure}{\imgwidth}
        \centering
        \includegraphics[width=\linewidth]{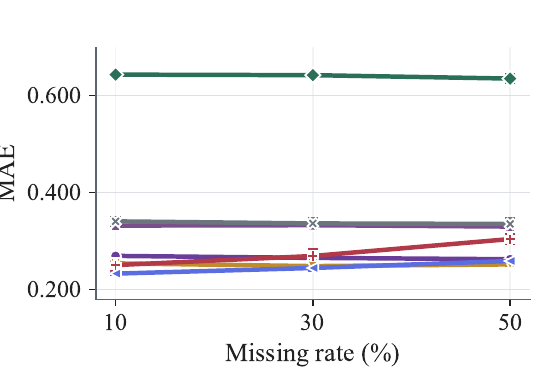}
        \caption{Default}
    \end{subfigure}\hfill
    \begin{subfigure}{\imgwidth}
        \centering
        \includegraphics[width=\linewidth]{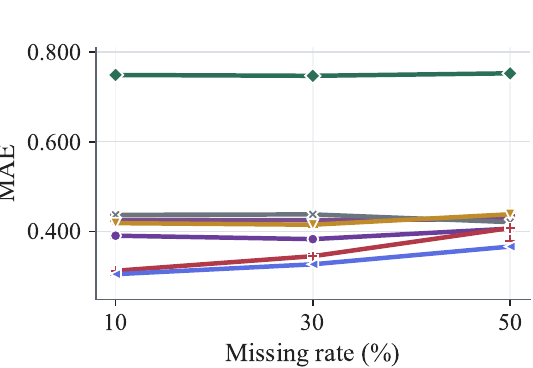}
        \caption{Gesture}
    \end{subfigure}
    \begin{subfigure}{\imgwidth}
        \centering
        \includegraphics[width=\linewidth]{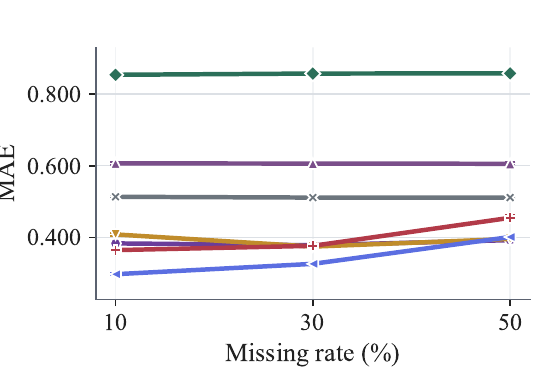}
        \caption{Letter}
    \end{subfigure}
    \begin{subfigure}{\imgwidth}
        \centering
        \includegraphics[width=\linewidth]{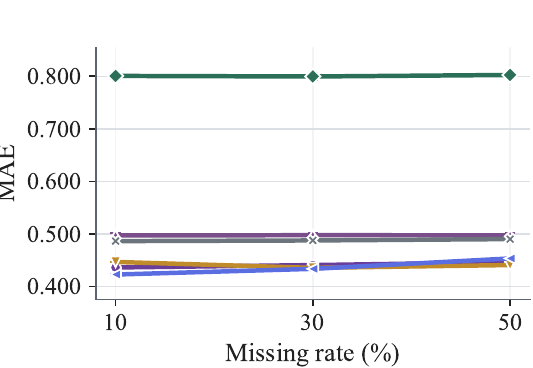}
        \caption{Magic}
    \end{subfigure}
    \caption{Per-dataset average test imputation MAE of MCAR missing data at a 50\% missing rate across ten missing masks. The horizontal axis indicates the training missing rate.}

    \label{fig:app_tab_mae}
\end{figure*}

\begin{figure*}[t]
    \centering
    \begin{subfigure}{\linewidth}
    \centering
    \includegraphics[width=\linewidth]{figs/legend.pdf}
    \end{subfigure}\hfill
    \def\imgwidth{0.33\textwidth}
    \begin{subfigure}{\imgwidth}
    \centering
    \includegraphics[width=\linewidth]{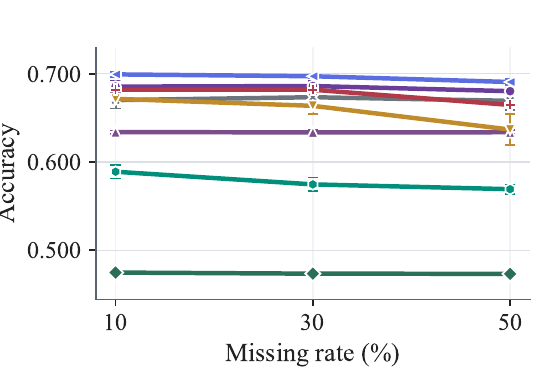}
    \caption{Adult}
    \end{subfigure}\hfill
    \begin{subfigure}{\imgwidth}
        \centering
    \includegraphics[width=\linewidth]{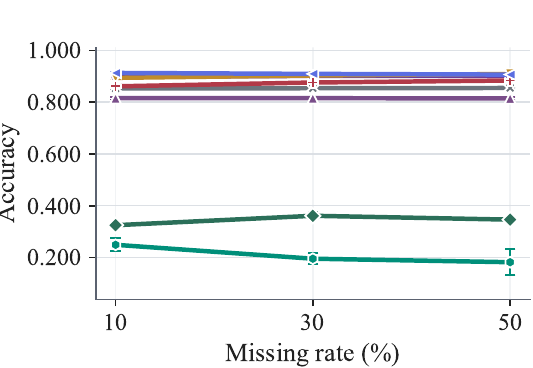}
        \caption{Bean}
    \end{subfigure}\hfill
    \begin{subfigure}{\imgwidth}
        \centering
        \includegraphics[width=\linewidth]{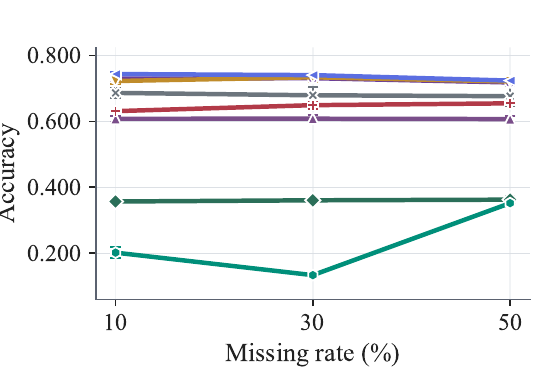}
        \caption{California}
    \end{subfigure}
    \begin{subfigure}{\imgwidth}
        \centering
        \includegraphics[width=\linewidth]{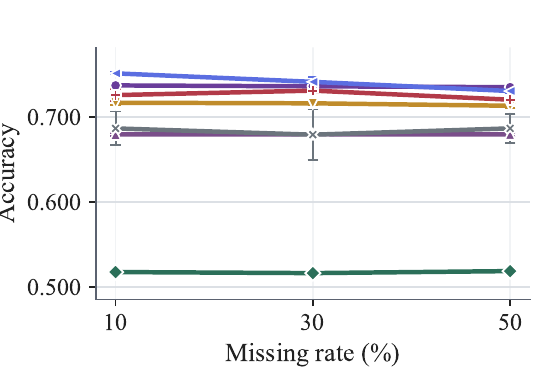}
        \caption{Default}
    \end{subfigure}\hfill
    \begin{subfigure}{\imgwidth}
        \centering
        \includegraphics[width=\linewidth]{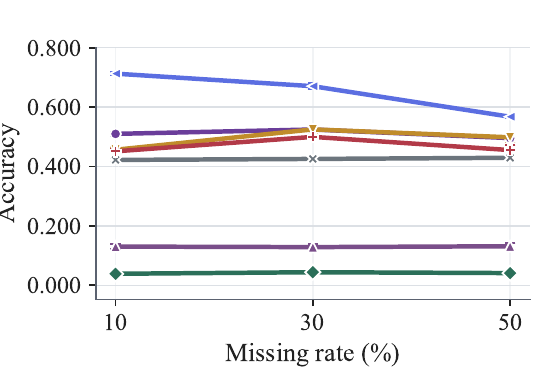}
        \caption{Letter}
    \end{subfigure}
    \begin{subfigure}{\imgwidth}
        \centering
        \includegraphics[width=\linewidth]{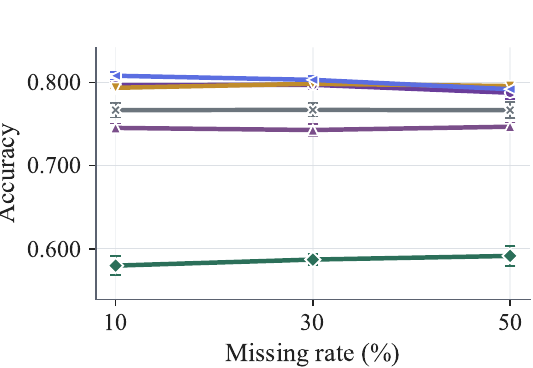}
        \caption{Magic}
    \end{subfigure}
    \caption{Per-dataset average test imputation accuracy of MCAR missing data at a 50\% missing rate across ten missing masks. The horizontal axis indicates the training missing rate.}
    \label{fig:app_tab_disc}
\end{figure*}

\begin{figure*}[t]
    \centering
    \begin{subfigure}{\linewidth}
    \centering
    \includegraphics[width=0.25\linewidth]{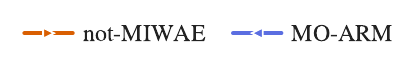}
    \end{subfigure}\hfill
    \def\imgwidth{0.24\textwidth}
    \begin{subfigure}{\imgwidth}
    \centering
    \includegraphics[width=\linewidth]{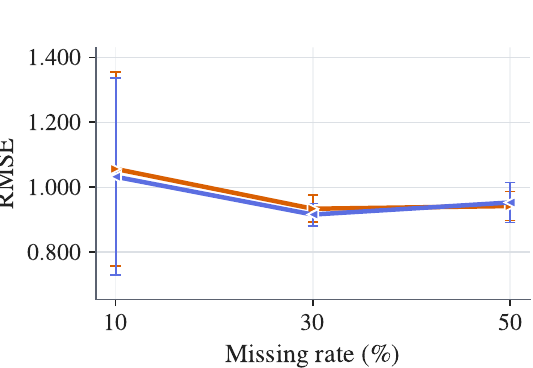}
    \caption{Adult}
    \end{subfigure}\hfill
    \begin{subfigure}{\imgwidth}
        \centering
    \includegraphics[width=\linewidth]{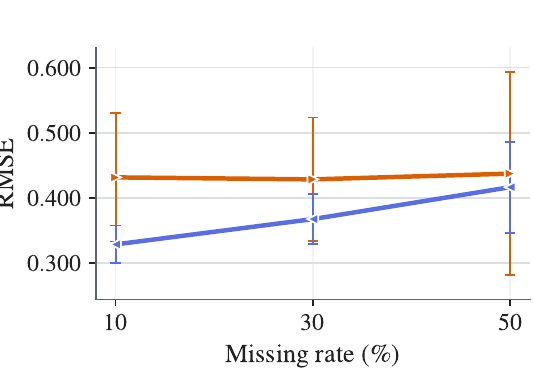}
        \caption{Bean}
    \end{subfigure}\hfill
    \begin{subfigure}{\imgwidth}
        \centering
        \includegraphics[width=\linewidth]{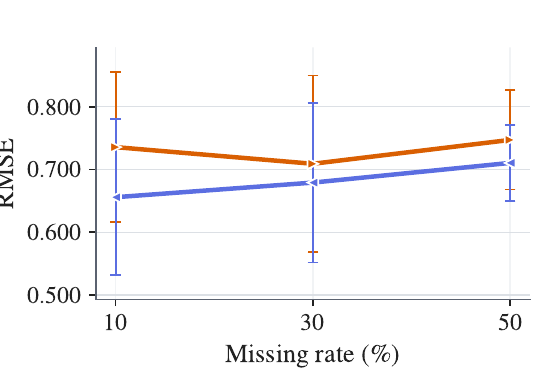}
        \caption{California}
    \end{subfigure}
    \begin{subfigure}{\imgwidth}
        \centering
        \includegraphics[width=\linewidth]{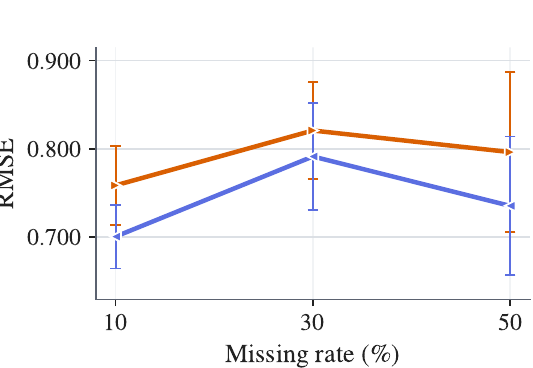}
        \caption{Default}
    \end{subfigure}\hfill
    \begin{subfigure}{\imgwidth}
        \centering
        \includegraphics[width=\linewidth]{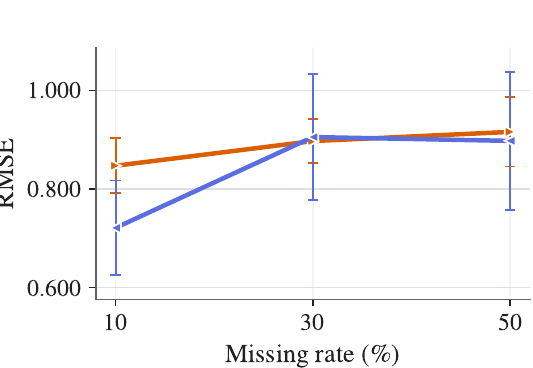}
        \caption{Gesture}
    \end{subfigure}
    \begin{subfigure}{\imgwidth}
        \centering
        \includegraphics[width=\linewidth]{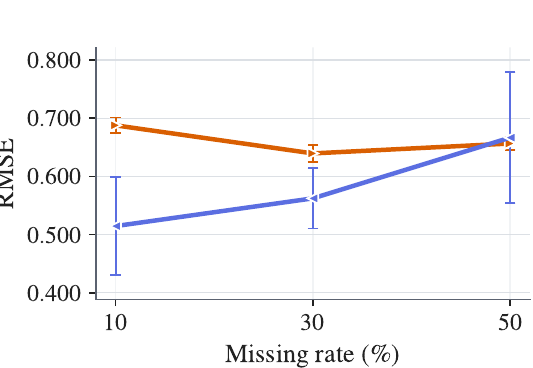}
        \caption{Letter}
    \end{subfigure}
    \begin{subfigure}{\imgwidth}
        \centering
        \includegraphics[width=\linewidth]{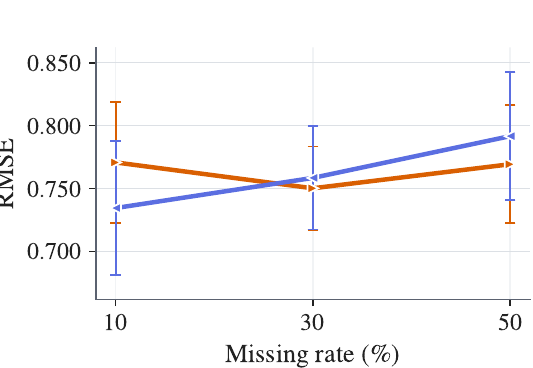}
        \caption{Magic}
    \end{subfigure}
    \caption{Per-dataset average test imputation RMSE of MNAR missing data at a 50\% missing rate across five missing masks. The horizontal axis indicates the training missing rate.}

    \label{fig:app_tab_MNAR_rmse}
\end{figure*}

\begin{figure*}[t]
    \centering
    \begin{subfigure}{\linewidth}
    \centering
    \includegraphics[width=0.25\linewidth]{figs/legend_MNAR.pdf}
    \end{subfigure}\hfill
    \def\imgwidth{0.24\textwidth}
    \begin{subfigure}{\imgwidth}
    \centering
    \includegraphics[width=\linewidth]{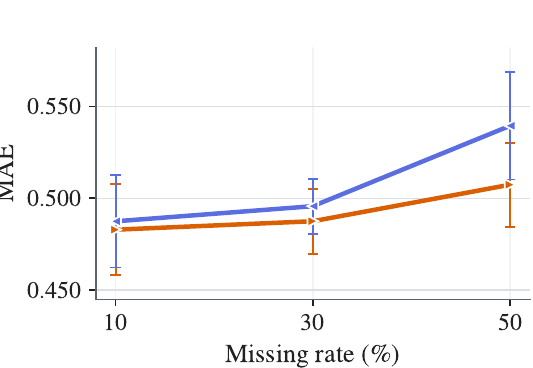}
    \caption{Adult}
    \end{subfigure}\hfill
    \begin{subfigure}{\imgwidth}
        \centering
    \includegraphics[width=\linewidth]{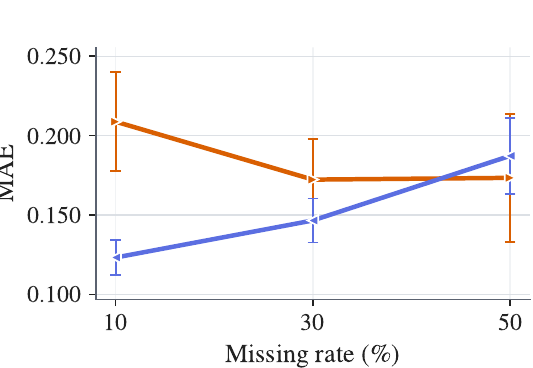}
        \caption{Bean}
    \end{subfigure}\hfill
    \begin{subfigure}{\imgwidth}
        \centering
        \includegraphics[width=\linewidth]{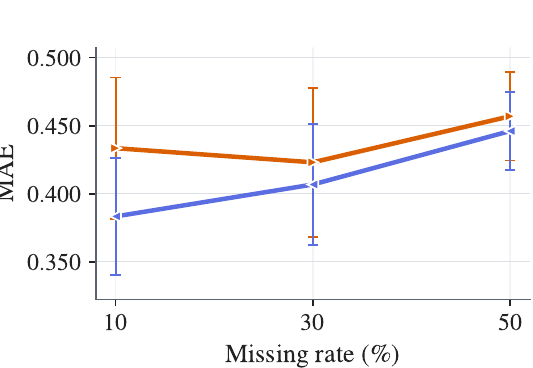}
        \caption{California}
    \end{subfigure}
    \begin{subfigure}{\imgwidth}
        \centering
        \includegraphics[width=\linewidth]{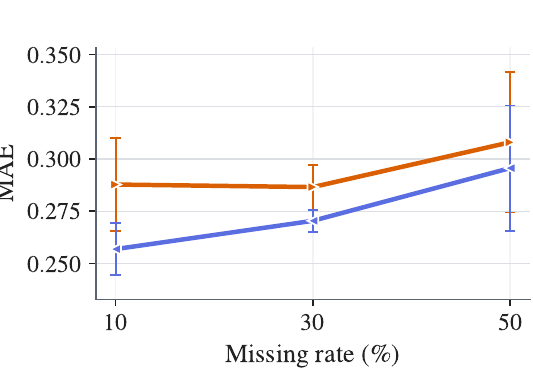}
        \caption{Default}
    \end{subfigure}\hfill
    \begin{subfigure}{\imgwidth}
        \centering
        \includegraphics[width=\linewidth]{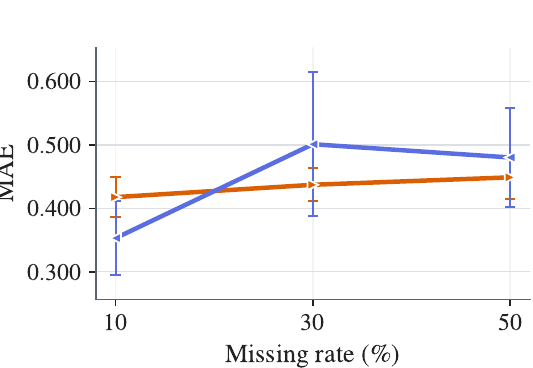}
        \caption{Gesture}
    \end{subfigure}
    \begin{subfigure}{\imgwidth}
        \centering
        \includegraphics[width=\linewidth]{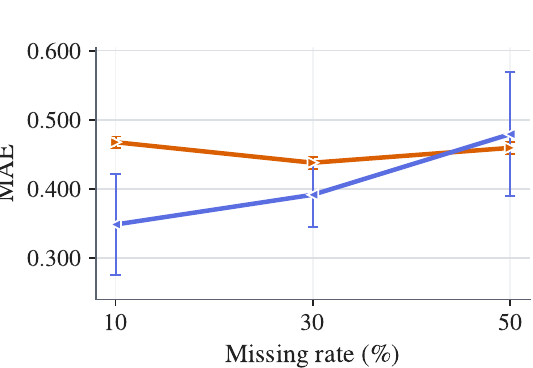}
        \caption{Letter}
    \end{subfigure}
    \begin{subfigure}{\imgwidth}
        \centering
        \includegraphics[width=\linewidth]{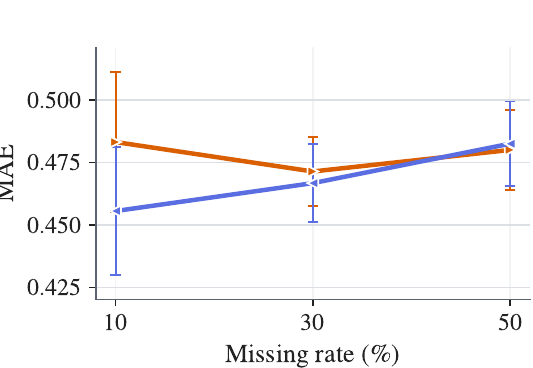}
        \caption{Magic}
    \end{subfigure}
    \caption{Per-dataset average test imputation MAE of MNAR missing data at a 50\% missing rate across ten missing masks. The horizontal axis indicates the training missing rate.}

    \label{fig:app_tab_MNAR_mae}
\end{figure*}

\begin{figure*}[t]
    \centering
    \begin{subfigure}{\linewidth}
    \centering
    \includegraphics[width=0.25\linewidth]{figs/legend_MNAR.pdf}
    \end{subfigure}\hfill
    \def\imgwidth{0.33\textwidth}
    \begin{subfigure}{\imgwidth}
    \centering
    \includegraphics[width=\linewidth]{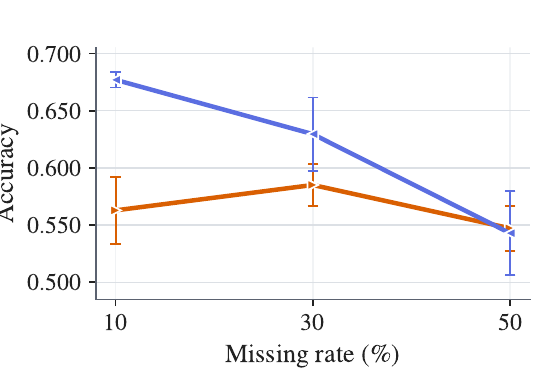}
    \caption{Adult}
    \end{subfigure}\hfill
    \begin{subfigure}{\imgwidth}
        \centering
    \includegraphics[width=\linewidth]{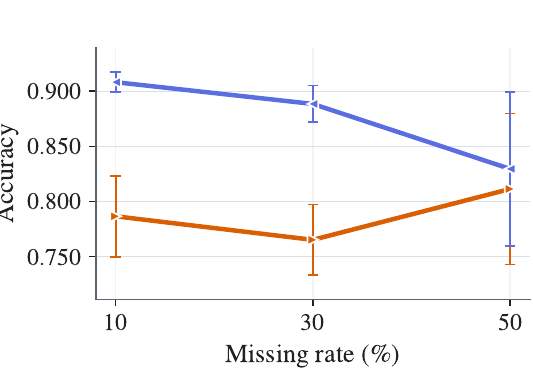}
        \caption{Bean}
    \end{subfigure}\hfill
    \begin{subfigure}{\imgwidth}
        \centering
        \includegraphics[width=\linewidth]{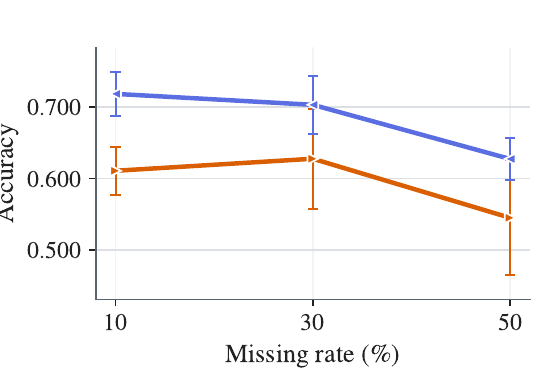}
        \caption{California}
    \end{subfigure}
    \begin{subfigure}{\imgwidth}
        \centering
        \includegraphics[width=\linewidth]{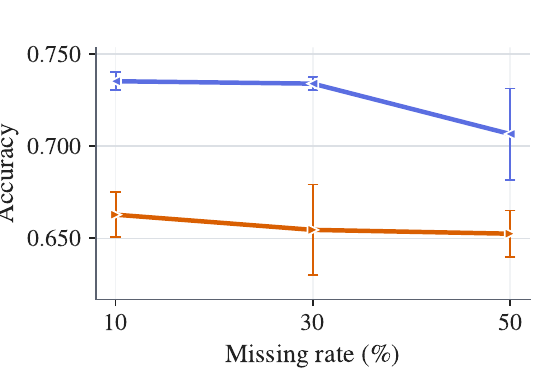}
        \caption{Default}
    \end{subfigure}\hfill
    \begin{subfigure}{\imgwidth}
        \centering
        \includegraphics[width=\linewidth]{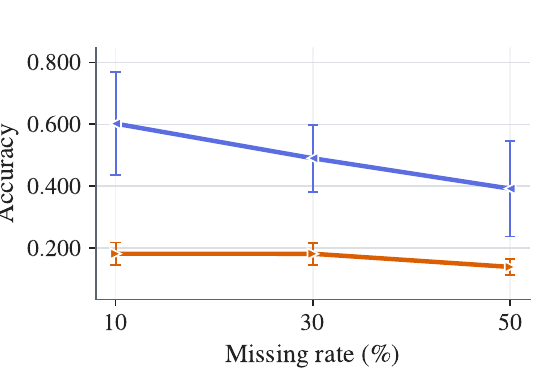}
        \caption{Letter}
    \end{subfigure}
    \begin{subfigure}{\imgwidth}
        \centering
        \includegraphics[width=\linewidth]{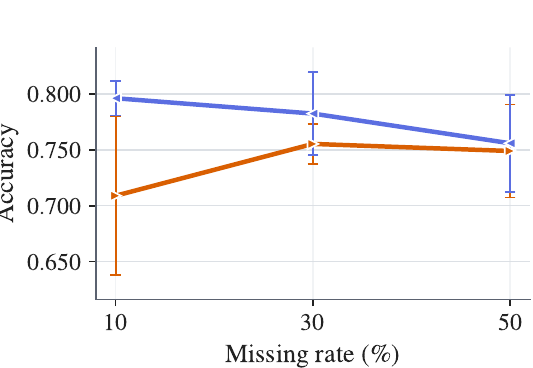}
        \caption{Magic}
    \end{subfigure}
    \caption{Per-dataset average test imputation accuracy of MNAR missing data at a 50\% missing rate across ten missing masks. The horizontal axis indicates the training missing rate.}
    \label{fig:app_tab_MNAR_disc}
\end{figure*}

We report here the per-dataset imputation results for all tabular benchmarks considered in the paper. For MCAR missingness, Figures~\ref{fig:app_tab_rmse} and~\ref{fig:app_tab_mae} summarize the performance on continuous variables in terms of RMSE and MAE, respectively, while Figure~\ref{fig:app_tab_disc} reports the accuracy on categorical variables. For MNAR missingness, the corresponding per-dataset results are shown in Figures~\ref{fig:app_tab_MNAR_rmse}, \ref{fig:app_tab_MNAR_mae}, and~\ref{fig:app_tab_MNAR_disc}.

\end{document}